\documentclass[preprint,12pt, authoryear]{elsarticle}
\usepackage[utf8x]{inputenc}
\usepackage[T1]{fontenc}
\usepackage{mathptmx} %
\usepackage{amssymb} %
\usepackage{amsmath}
\usepackage{calc} %
\usepackage{enumitem} %
\usepackage[authoryear]{natbib}
\usepackage[a4paper, lmargin=0.1666\paperwidth, rmargin=0.1666\paperwidth, tmargin=0.1111\paperheight, bmargin=0.1111\paperheight]{geometry} %
\usepackage[pdftex,linkcolor=blue,urlcolor=blue,pdfborder={0 0 0}]{hyperref} %
\usepackage{booktabs}
\usepackage{subcaption}
\usepackage{hyperref}
\hypersetup{hypertex=true,
    colorlinks=true,
    linkcolor=true,
    anchorcolor=true,
    citecolor=true}
\usepackage{pifont}
\usepackage{threeparttable}
\usepackage[all]{nowidow} %
\usepackage[protrusion=true,expansion=true]{microtype} %

\journal{Computers, Environment and Urban Systems}
\usepackage{url}
\usepackage{boxedminipage}
\usepackage{textpos}

\begin{document}

\begin{frontmatter}

\title{ZenSVI: An Open-Source Software for the Integrated Acquisition, Processing and Analysis of Street View Imagery Towards Scalable Urban Science}

\author[doa]{Koichi Ito}
\author[doa]{Yihan Zhu}
\author[doa]{Mahmoud Abdelrahman}
\author[doa]{Xiucheng Liang}
\author[doa]{Zicheng Fan}
\author[doa]{Yujun Hou}
\author[szu]{Tianhong Zhao}
\author[doa,cuhk]{Rui Ma}
\author[doa,tak]{Kunihiko Fujiwara}
\author[doa,wu]{Jiani Ouyang}
\author[sec]{Matias Quintana}
\author[doa,dre]{Filip Biljecki\corref{cor1}}

\cortext[cor1]{Corresponding author. Email: filip@nus.edu.sg}
\affiliation[doa]{organization={Department of Architecture, National University of Singapore}, 
            addressline={4 Architecture Drive}, 
            city={Singapore},
            postcode={117356}, 
            country={Singapore}}
\affiliation[szu]{organization={College of Big Data and Internet, Shenzhen Technology University},%
            addressline={3002 Lantian Road, Shenzhen}, 
            postcode={518118}, 
            country={China}}
\affiliation[cuhk]{organization={Department of Architecture and Civil Engineering, City University of Hong Kong},%
            addressline={Hong Kong SAR}, 
            postcode={999077}, 
            country={China}}
\affiliation[tak]{organization={Research \& Development Institute, Takenaka Corporation},%
            addressline={1-5-1 Otsuka, Inzai, Chiba}, 
            postcode={270-1352}, 
            country={Japan}}
\affiliation[wu]{organization={State Key Laboratory of Information Engineering in Surveying, Mapping and Remote Sensing, Wuhan University},%
            addressline={Wuhan}, 
            postcode={430079}, 
            country={China}}
\affiliation[sec]{organization={Singapore-ETH Centre, Future Cities Lab Global Programme},
            addressline={CREATE campus, 1 Create Way, \#06-01 CREATE Tower}, 
            postcode={138602}, 
            country={Singapore}}
\affiliation[dre]{organization={Department of Real Estate, National University of Singapore},%
            addressline={15 Kent Ridge Drive}, 
            city={Singapore},
            postcode={119245}, 
            country={Singapore}}     
            
\begin{abstract}
\begin{textblock*}{\textwidth}(-2cm,-16.6cm)
\begin{center}
\begin{footnotesize}
\begin{boxedminipage}{1.3\textwidth}
This is the Accepted Manuscript version of an article published by Elsevier in the journal \emph{Computers, Environment and Urban Systems} in 2025, which is available at: \url{https://doi.org/10.1016/j.compenvurbsys.2025.102283}\\ Cite as:
Ito K, Zhu Y, Abdelrahman M, Liang X, Fan Z, Hou Y, Zhao T, Ma R, Fujiwara K, Ouyang J, Quintana M, Biljecki F (2025): ZenSVI: An open-source software for the integrated acquisition, processing and analysis of street view imagery towards scalable urban science. \textit{Computers, Environment and Urban Systems}, 119: 102283.
\end{boxedminipage}
\end{footnotesize}
\end{center}
\end{textblock*}

\begin{textblock*}{1.5\textwidth}(-1.6cm,12.5cm)
{\tiny{\copyright{ }2025, Elsevier. Licensed under the Creative Commons Attribution-NonCommercial-NoDerivatives 4.0 International (\url{http://creativecommons.org/licenses/by-nc-nd/4.0/})}}
\end{textblock*}
Street view imagery (SVI) has been instrumental in many studies in the past decade to understand and characterize street features and the built environment. Researchers across a variety of domains, such as transportation, health, architecture, human perception, and infrastructure have employed different methods to analyze SVI. However, these applications and image-processing procedures have not been standardized, and solutions have been implemented in isolation, often making it difficult for others to reproduce existing work and carry out new research. Using SVI for research requires multiple technical steps: accessing APIs for scalable data collection, preprocessing images to standardize formats, implementing computer vision models for feature extraction, and conducting spatial analysis. These technical requirements create barriers for researchers in urban studies, particularly those without extensive programming experience. We developed ZenSVI, a free and open-source Python package that integrates and implements the entire process of SVI analysis, supporting a wide range of use cases. Its end-to-end pipeline includes downloading SVI from multiple platforms (e.g., Mapillary and KartaView) efficiently, analyzing metadata of SVI, applying computer vision models to extract target features, transforming SVI into different projections (e.g., fish-eye and perspective) and different formats (e.g., depth map and point cloud), visualizing analyses with maps and plots, and exporting outputs to other software tools. We demonstrated its use in Singapore through a case study of data quality assessment and clustering analysis in a streamlined manner. Our software improves the transparency, reproducibility, and scalability of research relying on SVI and supports researchers in conducting urban analyses efficiently. Its modular design facilitates extensions of the package for new use cases.
This package is openly available at \url{https://github.com/koito19960406/ZenSVI}, and it is supported by documentation including tutorials (\url{https://zensvi.readthedocs.io/en/latest/examples/index.html}).
\end{abstract}


\end{frontmatter}

\section{Introduction} \label{sec:intro}
Street view imagery (SVI) is a popular geospatial dataset in various fields pertaining to urban science, such as transportation, health, morphology, and urban visual perception, thanks to its scalability, coverage, and spatiotemporal granularity \citep{biljecki_street_2021, rzotkiewicz_systematic_2018, kang_review_2020, ito_understanding_2024, wang_mapping_2024}.
Machine learning techniques to process image data have enabled automated assessments of SVI on large scales, and such scalability has been leveraged with the growing coverage of SVI data platforms. 
Also importantly, SVI's ability to capture street-level information and its spatiotemporal granularity has supported many micro-scale assessments of streets and neighborhoods over time that were difficult or impossible with conventional spatial data, such as remote sensing imagery and field surveys~\citep{fan_urban_2023,kang_review_2020,wang_assessing_2024,liang_revealing_2023}. 

The strengths of SVI have enabled numerous applications in urban studies and data-driven planning. 
Researchers have developed various indices to quantify urban phenomena, from the green view index \citep{li_assessing_2015,yang_can_2009} to sky view and building view factors \citep{liang_automatic_2017,wang_assessing_2022,2023_jag_svi_sensitivity}, and comprehensive walkability and bikeability indices \citep{koo_how_2022,ito_assessing_2021}. 
These SVI-derived indicators have been used to analyze diverse urban phenomena including land use patterns, housing prices \citep{law_take_2019a,zhao_quantitative_2023a}, transportation flows \citep{jiao_forecasting_2023}, and socioeconomic status \citep{fan_urban_2023,ning_converting_2022,cao_integrating_2018,Hu2024-xu,Fujiwara2024-vs}. 
SVI has also been utilized to predict or estimate objective and subjective characteristics of streets, such as road condition assessment \citep{ai_automatic_2018,shu_pavement_2021,demesquita_street_2022,Arya2021}, building classification \citep{kang_building_2018,li_estimating_2018,sun2022understanding}, accessibility evaluation \citep{weld_deep_2019}, and assessment of subjective perceptions including safety \citep{kang2023assessing}, sound intensity \citep{zhao2023sensing}, architectural design \citep{liang2024evaluating}, and attractiveness \citep{dubey2016deep,zhang_measuring_2018}. 

Besides the achievements in maximizing usability, research relying on SVI owes its developments and popularity also to the availability of data. 
For example, major commercial providers (e.g., Google Street View) now have extensive coverage around the world, in particular, in North America, East, and Southeast Asia, and Europe.
Many researchers have taken advantage of such coverage, standardized data, and scalability to conduct large-scale studies comparing multiple cities, regions, and countries \citep{biljecki_street_2021, ito_examining_2024}.
The increasing popularity of crowdsourced SVI has also been a notable development --- anyone, anywhere can collect street-level imagery with consumer devices while moving and share it openly on platforms such as Mapillary, KartaView, and Panoramax.
This data stream has advantages such as free license, availability of additional points of view (e.g., imagery from cyclists), and spatiotemporal coverage that commercial services may not offer in some areas~\citep{2023_jag_svi_sensitivity, Helbich_2024, S_nchez_2024,hou2022comprehensive, Bendixen_2023,hou2024global}.

As much as SVI has been a popular and powerful instrument in sensing the built environment, there remain technical difficulties in conducting studies relying on it, especially for researchers without advanced analytical and technical skills. In fact, such difficulties have even been on the rise, as large-scale big data analyses with deep learning models are becoming the norm \citep{ito_understanding_2024}.
As a consequence, domain experts who are less technical find it challenging to conduct such studies, which hampers equity and participation in research \citep{ito_assessing_2021, bromm_virtual_2020, zhu2025understanding}, and may also impede the development of new use cases and the representation of certain geographical areas.
Further, most publications are not accompanied by openly released code to allow reproducibility, compounding the issue~\citep{biljecki_street_2021}.

This challenge has been common across many domains in urban analytics, including urban morphology. 
However, the growing adoption of open science practices in urban studies has begun to address this issue, with free and open-source software packages being developed to support various analytical tasks. 
These tools have democratized urban analytics by enabling researchers to conduct sophisticated analyses without requiring advanced technical expertise~\citep{yap_free_2022}.

OSMnx by \citet{boeing_osmnx_2017} is a quintessential example --- it is a Python package that has facilitated many applications that require network analysis or OpenStreetMap data by integrating them as a \textit{one-stop} software. 
Trackintel by \citet{martin_trackintel_2023} is also a Python package and provides many functions of GPS-trajectory data analysis with standardized models, thereby enhancing the transparency, transferability, and reproducibility of research. 
Another notable Python package is Urbanity developed by \citet{yap_urbanity_2023}, which allows users to download ready-to-use feature-rich network data. 
There have also been domain-specific tools, e.g.,\ an R package by \citet{mahajan_greenr_2024} computes the green index based on multiple sources of data, such as OpenStreetMap and SVI. 

As \citet{yap_free_2022} highlighted in their comprehensive systematic review, there is still a need for integrative open-source tools that can comparatively evaluate urban environments on a large scale and do so with relative ease of use and accessibility. 
This need is particularly acute in SVI research, where even though researchers occasionally share their code, these contributions often address specific niche problems, leading to a fragmented landscape of tools that are difficult to adapt or integrate into broader workflows. 
Despite the increasing number of applications, trends, and challenges mentioned above, researchers have only made a limited effort to standardize, facilitate, or integrate the workflow of SVI analysis, i.e., an integrated open-source solution to support urban studies relying on SVI.

Therefore, to bridge such a clear gap and facilitate reproducible SVI research, we introduce ZenSVI --- a framework and implementation that integrates downloading, processing, transforming, and visualizing SVI, which will benefit and encourage studies by improving the transparency, methodology, and ease of conducting research that requires SVI and associated elements such as computer vision models. 
This study is the first to create such an integrated software not only in SVI research but also in urban imagery.
ZenSVI simplifies many layers of SVI analysis, such as data acquisition and processing, and will therefore accelerate innovative urban science research by mitigating barriers for less experienced researchers and making their research reproducible. 
In a broader scope, its framework based on open data and open models presents a novel contribution to the open science current of urban analytics, and its modular framework may contribute to its extensions or integrations in other workflows.

The paper is organized as follows: Section \ref{sec:related_work} discusses existing studies, their research gaps, and our study's value, Section \ref{sec:framework} provides an overview of the package's framework, Section \ref{sec:case_study} shows how the package can be used in typical research settings, Section \ref{sec:extensibility_export} offers insights into how the package can further be extended or used by exporting data to other software applications, and lastly Section \ref{sec:discussion_conclusion} concludes this paper by highlighting the contribution of our work.

\section{Related work} \label{sec:related_work}
\subsection{Open-source software for urban science}
\begin{table}[htbp]
\centering
\caption{Comparison of urban analysis tools and libraries.}
\label{tab-opentools}
\begin{threeparttable}
\resizebox{\textwidth}{!}{
\begin{tabular}{lcccc}
\toprule
Title & Primary Application & Download & ML Applications & Visualization \\
\midrule
OSMnx \citep{boeing_osmnx_2017}\tnote{a} & Network Analysis & $\bullet$ & & $\bullet$ \\
Urbanity \citep{yap_urbanity_2023} \tnote{b} & Network Analysis & $\bullet$ & $\bullet$ & $\bullet$ \\
global-indicators \citep{Higgs2023} \tnote{c} & Mobility & $\bullet$ & & \\
PCT \citep{lovelace_propensity_2017} \tnote{d} & Mobility & $\bullet$ & & $\bullet$ \\
MovingPandas \citep{graser_movingpandas_2019} \tnote{e} & Mobility & & & $\bullet$ \\
scikit-mobility \citep{JSSv103i04} \tnote{f} & Mobility & & & $\bullet$ \\
rpi-urban-mobility-tracker \citep{Wojke2017simple, Wojke2018deep} \tnote{g} & Mobility & $\bullet$ & $\bullet$ & \\
Trackintel \citep{martin2023trackintel} \tnote{h} & Mobility & $\bullet$ & & $\bullet$ \\
GWmodelS \citep{lu2023gwmodels} \tnote{i} & Spatial analysis & $\bullet$ & & $\bullet$ \\
raster vision\tnote{j} & Satellite imagery & & $\bullet$ & \\
\bottomrule
\end{tabular}
}
\begin{tablenotes}
\tiny
\item[a] \url{https://github.com/gboeing/osmnx}
\item[b] \url{https://github.com/winstonyym/urbanity}
\item[c] \url{https://github.com/healthysustainablecities/global-indicators}
\item[d] \url{https://github.com/ITSLeeds/pct}
\item[e] \url{https://github.com/movingpandas/movingpandas}
\item[f] \url{https://github.com/scikit-mobility/scikit-mobility}
\item[g] \url{https://github.com/nathanrooy/rpi-urban-mobility-tracker}
\item[h] \url{https://github.com/mie-lab/trackintel}
\item[i] \url{https://github.com/GWmodel-Lab/GWmodelS}
\item[j] \url{https://github.com/azavea/raster-vision}
\end{tablenotes}
\end{threeparttable}
\end{table}

Open-source software tools are increasingly used in urban studies, enabling researchers to analyze complex urban phenomena efficiently and collaboratively \citep{yap_free_2022,Ponkanen2025}. 
These tools have contributed to the advancement of urban science by providing freely and easily accessible platforms for acquiring, processing, analyzing, and visualizing urban data. 
For example, OSMnx has facilitated the study of urban networks by streamlining the extraction, construction, and analysis of street networks from OpenStreetMap. 
Such tools allow researchers to tackle large-scale spatial analyses with minimal technical barriers.

The strength of these open-source tools lies not in introducing new scientific methods, but rather in their ability to democratize existing methodologies and foster collaborative improvement through community engagement. 
Their success stems from implementing established scientific approaches while remaining open to enhancement by a community of like-minded developers and researchers. 
These tools excel at facilitating essential tasks, such as downloading large datasets and visualizing complex spatial information, often serving as comprehensive one-stop solutions for specific analytical needs. 
Additionally, some have integrated machine learning (ML) approaches for tasks such as feature extraction without the need to build models from scratch (Table~\ref{tab-opentools}). 
By embedding cutting-edge algorithms and workflows in user-friendly interfaces, these tools empower researchers across disciplines, including those without extensive programming experience, to conduct innovative studies. 
This inclusivity expands the reach and impact of urban science.

\subsection{Open-source software for street view imagery analysis}

SVI has a wide range of applications in urban studies \citep{biljecki_street_2021}, including urban planning and management \citep{wang2023diagnosis,li2022urban,xia2021development,lu2019using,maniat2021deep,wang_assessing_2022}, environmental monitoring and evaluation \citep{cao_integrating_2018,he2021urban,xia2021development,li2022measuring,yin2016measuring,chen2023automatic,li2022associations,liang_revealing_2023,lu2023assessing,sanusi2016street,li2018quantifying}, transportation research \citep{campbell2019detecting,cai2022applying}, and socioeconomic studies \citep{alhasoun2019streetify,zhang2018impacts}. 
Although these applications solve specific problems, they share many of the same approaches and data sources, but no effort has been made to integrate these applications into one platform so that other researchers can reproduce their work easily.

SVI data have diverse sources and acquisition methods. 
The most widely used SVI comes from commercial GIS service platforms such as Google\footnote{\url{https://www.google.com/maps}}, Amap\footnote{\url{https://www.amap.com/}} and Baidu\footnote{\url{https://map.baidu.com/}}, which capture high-resolution geotagged panoramic images via vehicles \citep{mahabir2020crowdsourcing, chen2019evaluating}. 
These platforms offer a wide coverage suitable for urban research but are subject to terms of use and privacy policies \citep{hou2022comprehensive, Helbich_2024}. 
Open projects like KartaView and Mapillary consolidate and provide user-uploaded SVI, offering free, scalable alternatives, although often with lower consistency and quality \citep{hara2013combining,zheng2023method,juhasz2016user,hou2024global}. 
Additionally, some city governments and public institutions collect and release SVI, and social media platforms (e.g., Flickr and Instagram) are also significant sources of imagery \citep{bahrehdar2020streets,chen2020quantifying,callau2019landscape}.

Processing SVI typically involves complex computer vision techniques to extract and analyze valuable information, broadly categorized into traditional methods and deep learning-based approaches. 
The former includes edge detection, texture analysis, color feature extraction, and shape feature extraction \citep{bansal2011geo, de2015design}.
With advancements in deep learning, object detection, semantic segmentation, and image classification have become predominant. 
Object detection identifies and locates objects (e.g., vehicles, pedestrians, traffic signs) using models such as YOLO, Faster R-CNN, and R-CNN \citep{krylov2018automatic,campbell2019detecting,kang_review_2020, ren2023yolov5s,zhang2020quantifying}. 
Semantic segmentation assigns pixel-level labels (e.g., roads, buildings, vegetation) via models such as U-Net, SegNet, and DeepLab~\citep{xia2021sky,aikoh2023comparing,wang2022numerical, gou2022study,zhao2023sensing}. 
Image classification categorizes entire images (e.g., commercial, residential, industrial) using models such as ResNet and VGG \citep{law2020street,fang2022spatial,huang2024estimating}.

SVI data processing is a multistage process, with each stage requiring specific libraries. 
The initial stage of SVI data processing involves the downloading of data, which typically entails extracting and downloading high-quality SVI from online services such as the Google Street View API.
Traditional computer vision processing libraries such as OpenCV\footnote{\url{https://opencv.org/}} and Pillow\footnote{\url{https://pypi.org/project/pillow/}} are widely used for processing these images. 
OpenCV offers a rich set of image processing tools for tasks such as cropping, scaling, and rotation, and excels in computer vision tasks such as edge detection and feature extraction. 
Pillow, on the other hand, is an easy-to-use image processing library suitable for image format conversion, basic image processing, and batch image operations. 
As processing demands increase, deep learning processing libraries become essential. These libraries can perform complex tasks such as semantic segmentation and scene recognition. 
TensorFlow\footnote{\url{https://www.tensorflow.org/}} and PyTorch\footnote{\url{https://pytorch.org/}} are currently the most commonly used deep learning frameworks. 
They support the construction and training of deep neural networks, effectively handling large amounts of SVI data. Through these libraries, advanced tasks such as object detection and semantic segmentation can be achieved, allowing for a deeper understanding of the content within SVI.

In addition to conventional image processing and deep learning libraries, there are specialized libraries designed for SVI data processing and customized functionalities. These libraries often incorporate deep learning applications adapted from existing studies.
\autoref{tab_lib} summarizes common libraries that target SVI data. 
In terms of programming languages, the primary languages are Python, C++, and R, with some libraries having developed QGIS plugins. 
Some plugins support a pipeline of functions from data downloading to data processing. 
For example, \citet{li_assessing_2015} developed Treepedia, which uses computer vision techniques applied to Google Street View images to calculate the GVI for measuring and mapping vegetation coverage along urban streets. 
It establishes a workflow encompassing sampling point construction, SVI acquisition from Google Street View, metric calculation, and result visualization.

\begin{table}[]
\centering
\caption{Comparison of SVI data processing libraries.}
\label{tab_lib}
\resizebox{\textwidth}{!}{
\begin{threeparttable}
\begin{tabular}{@{}llllllllll@{}}
\toprule
Package         & \begin{tabular}[c]{@{}l@{}}Platform \&\\ language\end{tabular} & Download & \begin{tabular}[c]{@{}l@{}}Quality \\ assessment\end{tabular} & \begin{tabular}[c]{@{}l@{}}Metadata \\ extraction\end{tabular} & \begin{tabular}[c]{@{}l@{}}Image \\ transformation\end{tabular} & \begin{tabular}[c]{@{}l@{}}Deep \\ learning\end{tabular} &  Application               &Data source              \\ \midrule
streetlearn~\citep{mirowski2018learning}\tnote{a}          & C++, Python           & \ding{51}        & \ding{53}&  \ding{51}& \ding{53}& \ding{51}& Navigation                &Google                  \\
street-view-pipeline\tnote{b} & Python               & \ding{51}        & \ding{53}&  \ding{53}& \ding{53}& \ding{53}& Calculating GVI             &Google, OSM              \\
streetview\tnote{c}           & Python               & \ding{51}        & \ding{53}&  \ding{51}& \ding{53}& \ding{53}& Retrieving Google Street View            &Google                  \\
CODING~\citep{9380541}\tnote{d}               & Python               & \ding{53}        & \ding{53}&  \ding{51}& \ding{53}& \ding{51}& Building detection        &Google                  \\
urban-perceptions~\citep{muller2022city}\tnote{e}    & Python, QGIS          & \ding{51}        & \ding{53}&  \ding{53}& \ding{51}& \ding{51}& Street quality assessment &Google, Mapillary, others \\
Treepedia\tnote{f}            & Python               & \ding{51}        & \ding{53}&  \ding{51}& \ding{53}& \ding{53}& Calculating GVI             &Google                  \\
Green View Index\tnote{g}     & Python, QGIS          & \ding{51}        & \ding{53}&  \ding{51}& \ding{51}& \ding{53}& Calculating GVI             &Google                  \\
go2mapillary\tnote{h}         & Python, QGIS          & \ding{51}        & \ding{53}&  \ding{51}& \ding{53}& \ding{53}& Metadata extraction       &Mapillary               \\
googleway\tnote{i}            & R                    & \ding{51}        & \ding{53}&  \ding{51}& \ding{53}& \ding{53}& Accessing and plotting    &Google                  \\
StreetView-NatureVisibility\tnote{j} & Python          & \ding{51}        & \ding{51}&  \ding{53}& \ding{51}& \ding{51}& Calculating GVI             &Google                  \\
percept~\citep{danish_citizen_2025}\tnote{k}              & Python               & \ding{51}        & \ding{51}&  \ding{51}& \ding{51}& \ding{53}& Survey for human perception &Mapillary               \\
streetscape~\citep{yang2025streetscape}\tnote{l}              & R               & \ding{51}        & \ding{51}&  \ding{51}& \ding{53}& \ding{51}& GVI and human perception &Mapillary               \\\midrule
\textbf{ZenSVI}\tnote{m}\hspace*{1em} (\textit{Our work})              & Python               & \ding{51}        & \ding{51}&  \ding{51}& \ding{51}& \ding{51}& \begin{tabular}[c]{@{}l@{}}Acquisition, processing\\and analysis of SVI\end{tabular} & \begin{tabular}[c]{@{}l@{}}Mapillary, KartaView,\\Amsterdam SVI\end{tabular} \\ \bottomrule
\end{tabular}
\begin{tablenotes}
\item[a] \url{https://github.com/google-deepmind/streetlearn}
\item[b] \url{https://github.com/datasciencecampus/street-view-pipeline}
\item[c] \url{https://github.com/robolyst/streetview}
\item[d] \url{https://github.com/kyle-one/Context-Encoding-of-Detected-Buildings}
\item[e] \url{https://github.com/emilymuller1991/urban-perceptions}
\item[f] \url{https://github.com/mittrees/Treepedia_Public}
\item[g] \url{https://github.com/kowalski93/Green-View-Index-for-QGIS}
\item[h] \url{https://github.com/enricofer/go2mapillary}
\item[i] \url{https://github.com/SymbolixAU/googleway}
\item[j] \url{https://github.com/Spatial-Data-Science-and-GEO-AI-Lab/StreetView-NatureVisibility}
\item[k] \url{https://github.com/Spatial-Data-Science-and-GEO-AI-Lab/percept}
\item[l] \url{https://github.com/land-info-lab/streetscape}
\item[m] \url{https://github.com/koito19960406/ZenSVI}
\end{tablenotes}
\end{threeparttable}
}
\end{table}

Different libraries support various applications such as data exploration or metadata acquisition, green view index calculation, street quality assessment, and building feature exploration. 
The methods for computing street view features include traditional computer vision techniques (e.g., edge detection, grayscale calculation) and deep learning-based methods (e.g., scene classification, semantic segmentation). 
Most libraries use Google Street View data as their primary data source, but some libraries support the import of other data sources. 
For instance, the `urban-perceptions' and `percept' libraries support not only Google Street View data but also other sources such as Mapillary \citep{muller2022city, danish_citizen_2025}.
Prior to analysis, data cleaning is essential to ensure the usability of SVI. 
This involves the provision of metadata, quality assessment to determine whether images are suitable for analysis, and image transformation to convert a panoramic image into a regular image.

However, the existing specialized libraries are typically designed for single-purpose tasks, there is still a lack of a one-stop solution that provides unified support for downloading, cleaning, and multi-purpose SVI analytics, supporting a wide range of use cases.
Our integrated Python package aspires to contribute to many domains by filling this gap.

\section{ZenSVI framework and implementation} \label{sec:framework}
ZenSVI is conceived as a one-stop open-source solution for SVI analytics and is built with a modular architecture to ensure maintainability, extensibility, and code quality. 
The package is developed with comprehensive unit tests, automated continuous integration/continuous deployment (CI/CD) pipelines through GitHub Actions and detailed API documentation with usage examples. 
Further, as an open-source project, ZenSVI welcomes contributions from the community, enabling researchers and developers to extend its functionality, improve existing features, or address specific use cases. 
This continuous and collaborative development approach ensures that the package can evolve alongside the changing needs of the urban research community while maintaining high standards of code quality and documentation.

The framework is designed with compatibility in mind, leveraging widely used Python packages in the geospatial ecosystem. 
Core dependencies include GeoPandas\footnote{\url{https://github.com/geopandas/geopandas}} for spatial operations~\citep{kelsey_jordahl_2020_3946761}, OSMnx for street network analysis, and PyTorch for deep learning implementations. 
This foundation ensures seamless integration with existing geospatial workflows while maintaining forward compatibility through semantic versioning. 
The framework's reliance on these well-maintained, community-supported packages helps ensure long-term continuity and stability. 

For computer vision capabilities, we integrated established models from other research efforts, maintaining their original architectures and parameters to ensure reproducibility and validated performance. 
Also, each implemented model is documented with citations to its source publication.

\begin{figure}[tbp]
    \includegraphics[width=.95\textwidth]{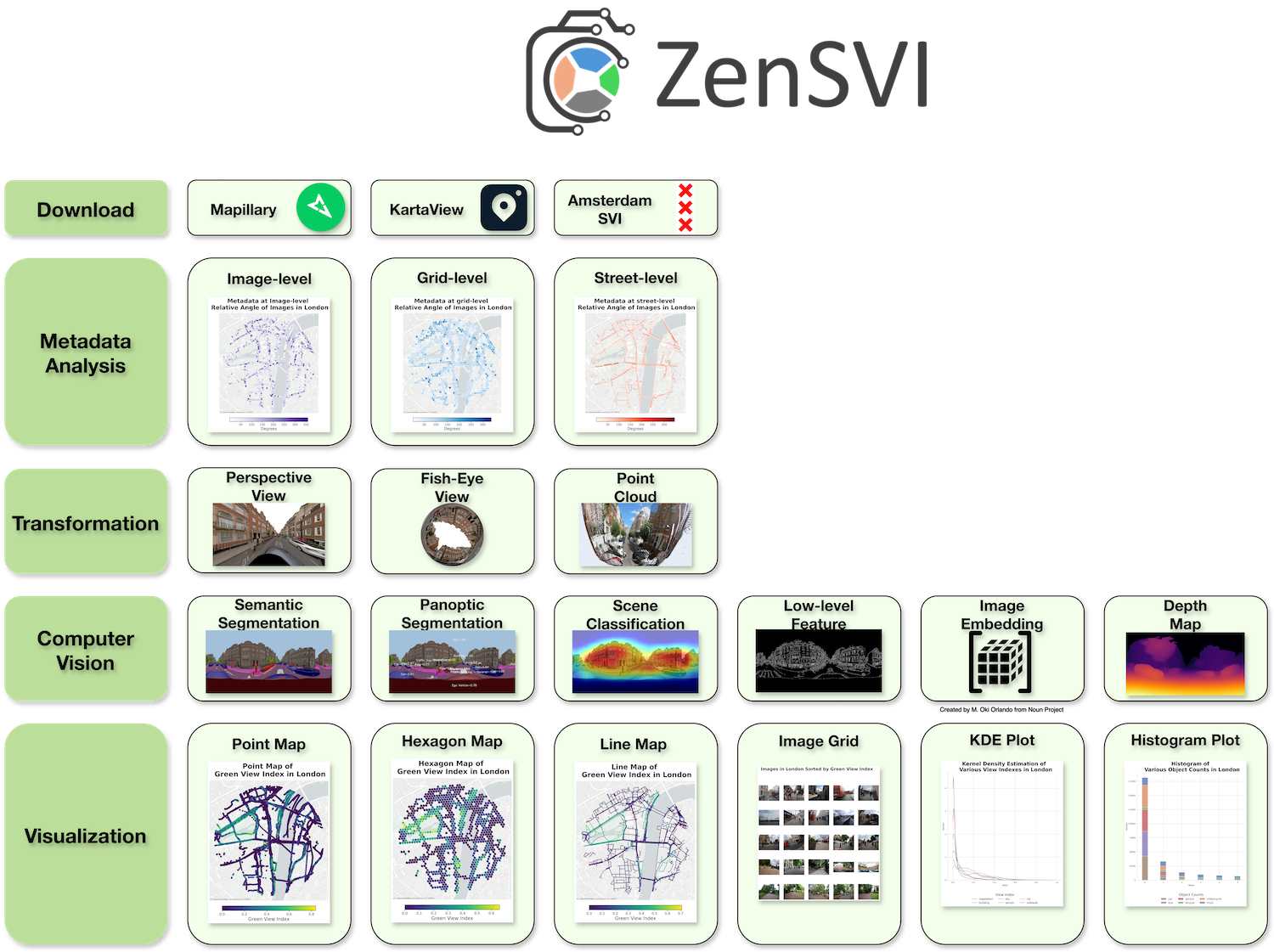}
    \caption{Overview of the framework of ZenSVI package, consisting of five sub-packages: 1) download, 2) metadata analysis, 3) computer vision, 4) image transformation, and 5) visualization.}
    \label{fig:framework:overview}
\end{figure}

The package architecture consists of five main components (\autoref{fig:framework:overview}), which are described separately in the subsequent sections:
\begin{itemize}
    \item Download --- Acquisition of imagery from multiple sources (Section~\ref{sc:download}).
    \item Metadata Analysis --- Computation of metadata based on spatiotemporal information at different spatial levels (Section~\ref{sc:metadata}).
    \item Transformation --- Transformation of images' projections and formats (Section~\ref{sc:transformation}).
    \item Computer Vision --- Multi-faceted analysis through traditional and deep learning techniques (Section~\ref{sc:cv}).
    \item Visualization --- Presentation of results through various maps and plot types (Section~\ref{sc:visualization}).
\end{itemize}

\subsection{Download}\label{sc:download}
Built on geospatial libraries such as GeoPandas and OSMnx \citep{boeing_osmnx_2017}, this sub-package implements wrapper functions with parallel processing to acquire SVI from two major platforms: Mapillary and KartaView, which have global coverage.
In addition, the City of Amsterdam's SVI API is supported, which showcases ZenSVI's capability to accommodate various data sources, not limited to commercial or crowdsourced platforms.

The package standardizes input types and output formats while enhancing the official APIs' functionalities. 
Users can provide location data for SVI acquisition in various formats: coordinate pairs, CSV files with longitude and latitude columns, shapefiles, GeoJSON files, or place names as string values. 
By leveraging GeoPandas for spatial operations and OSMnx for street network analysis, the package efficiently and robustly processes these inputs to generate and filter sampling points for SVI downloads with parallel processing and checkpoints. 
With these additional features, ZenSVI eliminates the need for custom API integration code by researchers.

\subsection{Metadata Analysis}\label{sc:metadata}
Metadata is important in facilitating exploratory data analyses and for adding more context to each image. 
This is particularly important for crowdsourced SVI because they are usually captured in diverse ways, unlike those from commercial platforms. 
Such inconsistency makes it difficult for researchers to know if a specific image is suitable for their research when they acquire numerous SVI (e.g., millions at the city scale). 
For example, if one is interested in nighttime walkability, it is unsuitable to use SVI from daytime or SVI captured on highways without sidewalks \citep{hou2024global}.

\begin{figure}[tbp]
    \includegraphics[width=\textwidth]{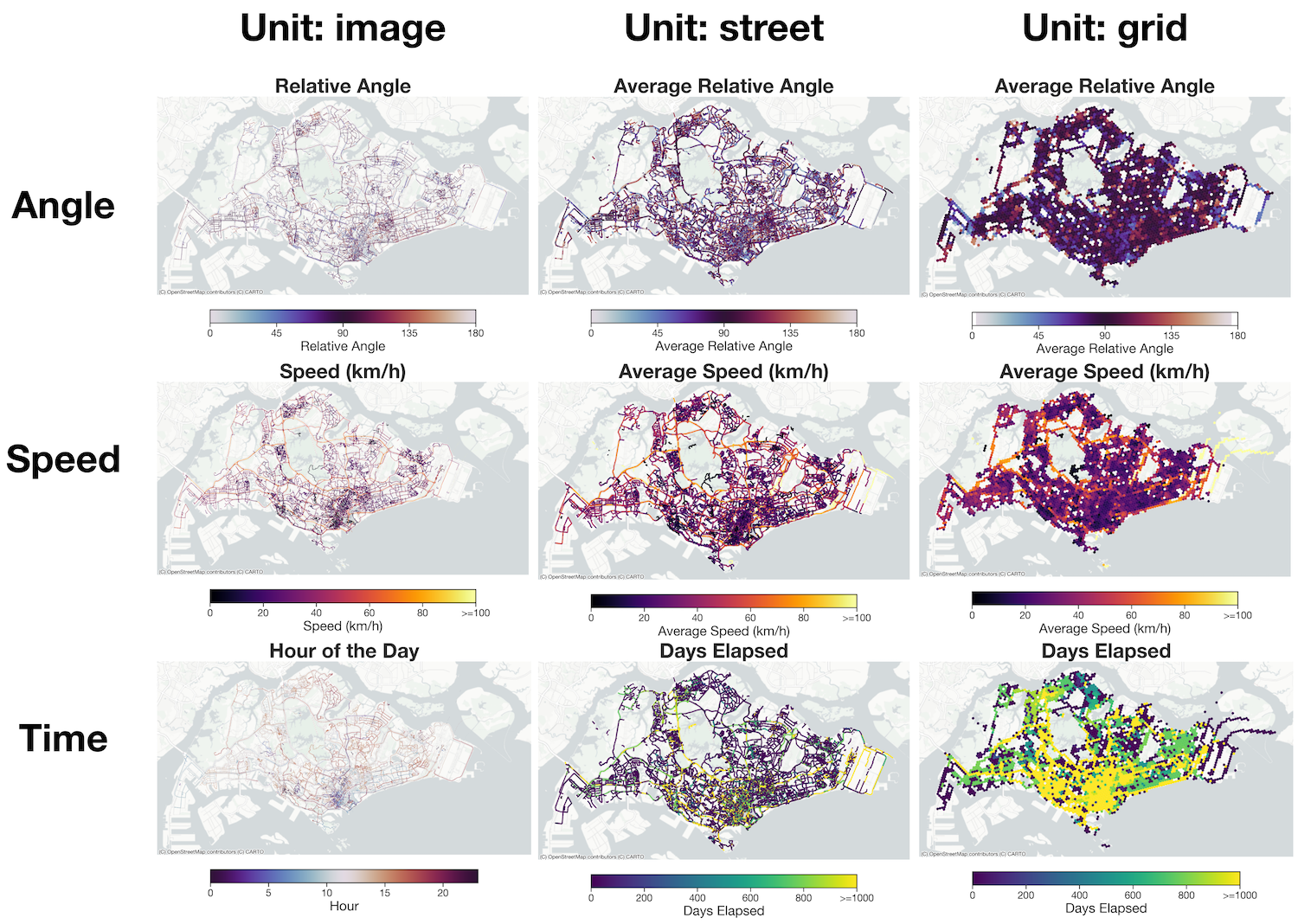}
    \caption{A matrix of maps representing various metadata. Columns show three different types of units of analysis: 1) image, 2) street, and 3) grid, and rows display three exemplary types of metadata: 1) angle, 2) speed, and 3) time.}
    \label{fig:framework:metadata}
\end{figure}

As exemplified in \autoref{fig:framework:metadata}, this sub-package for metadata analysis is built upon the basic metadata provided by Mapillary (such as capture time, camera parameters, and spatial coordinates) and further augments it by computing additional features. 
The package provides solutions for comprehensive metadata analysis at three levels: 1) image-level, 2) grid-level, and 3) street-level analyses.

At the most granular level, users can compute the indicators listed in \autoref{framework:tab:meta_image} for each of the SVI points (i.e., spatial locations of SVI) by inputting the CSV file outputted from the downloading sub-package after downloading SVI. 
These indicators extend beyond the basic Mapillary metadata to provide more detailed insights about each image.

For the street- and grid-level analyses, we match each SVI point with the nearest street and the corresponding grid cell to compute indicators at the aggregated level (see \autoref{framework:tab:meta_grid_street}), which are commonly used in use cases (e.g., average sky view factor at the district scale) and could also be useful for data quality analyses such as spatial and temporal coverage \citep{hou2022comprehensive}. 
This sub-package also contributes to the standardization of data properties by enabling researchers to explicitly query SVI based on both the original Mapillary metadata and our augmented indicators, allowing them to fetch only images suitable for their needs.

\begin{table}[tbp]
\centering
\caption{A list of metadata indicators at the image level.}
\footnotesize
\begin{tabular}{ll}
\toprule
\textbf{Attribute} & \textbf{Description} \\
\midrule
year & Year of the image collection. \\
month & Month of the image collection. \\
day & Day of the image collection. \\
hour & Hour of the image collection. \\
day\_of\_week & Day of the week of the image collection. \\
daytime\_nighttime & Daytime or nighttime based on astral package. \\
season & Season determined by hemisphere and month. \\
relative\_angle & Camera angle relative to nearest street segment. \\
h3\_id & Uber's H3 hexagon IDs (zoom level 0-15). \\
speed\_kmh & Speed in km/h, computed between points in a sequence. \\ \bottomrule
\end{tabular}
\label{framework:tab:meta_image}
\end{table}

\begin{table}[tbp]
\centering
\caption{A list of metadata indicators at the grid and street level.}
\footnotesize
\begin{tabular}{lp{7.5cm}}
\toprule
\textbf{Attribute} & \textbf{Description} \\
\midrule
coverage & Coverage of user-defined buffers from SVI points in percentage. \\
count & Number of SVI points. \\
days\_elapsed & Days between the oldest and newest SVI. \\
most\_recent\_date & Most recent date of SVI. \\
oldest\_date & Oldest date of SVI. \\
number\_of\_years & Number of unique years. \\
number\_of\_months & Number of unique months. \\
number\_of\_days & Number of unique days. \\
number\_of\_hours & Number of unique hours. \\
number\_of\_days\_of\_week & Number of unique days of the week. \\
number\_of\_daytime & SVIs collected during daytime. \\
number\_of\_nighttime & SVIs collected during nighttime. \\
number\_of\_spring & SVIs collected during spring. \\
number\_of\_summer & SVIs collected during summer. \\
number\_of\_autumn & SVIs collected during autumn. \\
number\_of\_winter & SVIs collected during winter. \\
average\_compass\_angle & Average compass angle of SVI. \\
average\_relative\_angle & Average relative angle to nearest street segment. \\
average\_is\_pano & Average ratio of panoramic SVIs. \\
number\_of\_users & Number of unique users contributing to SVI. \\
number\_of\_sequences & Number of unique sequences containing SVI. \\
number\_of\_organizations & Number of unique organizations contributing. \\
average\_speed\_kmh & Average speed in km/h. \\
\bottomrule
\end{tabular}
\label{framework:tab:meta_grid_street}
\end{table}

\subsection{Transformation}\label{sc:transformation}

\begin{figure}[tbp]
    \includegraphics[width=\textwidth]{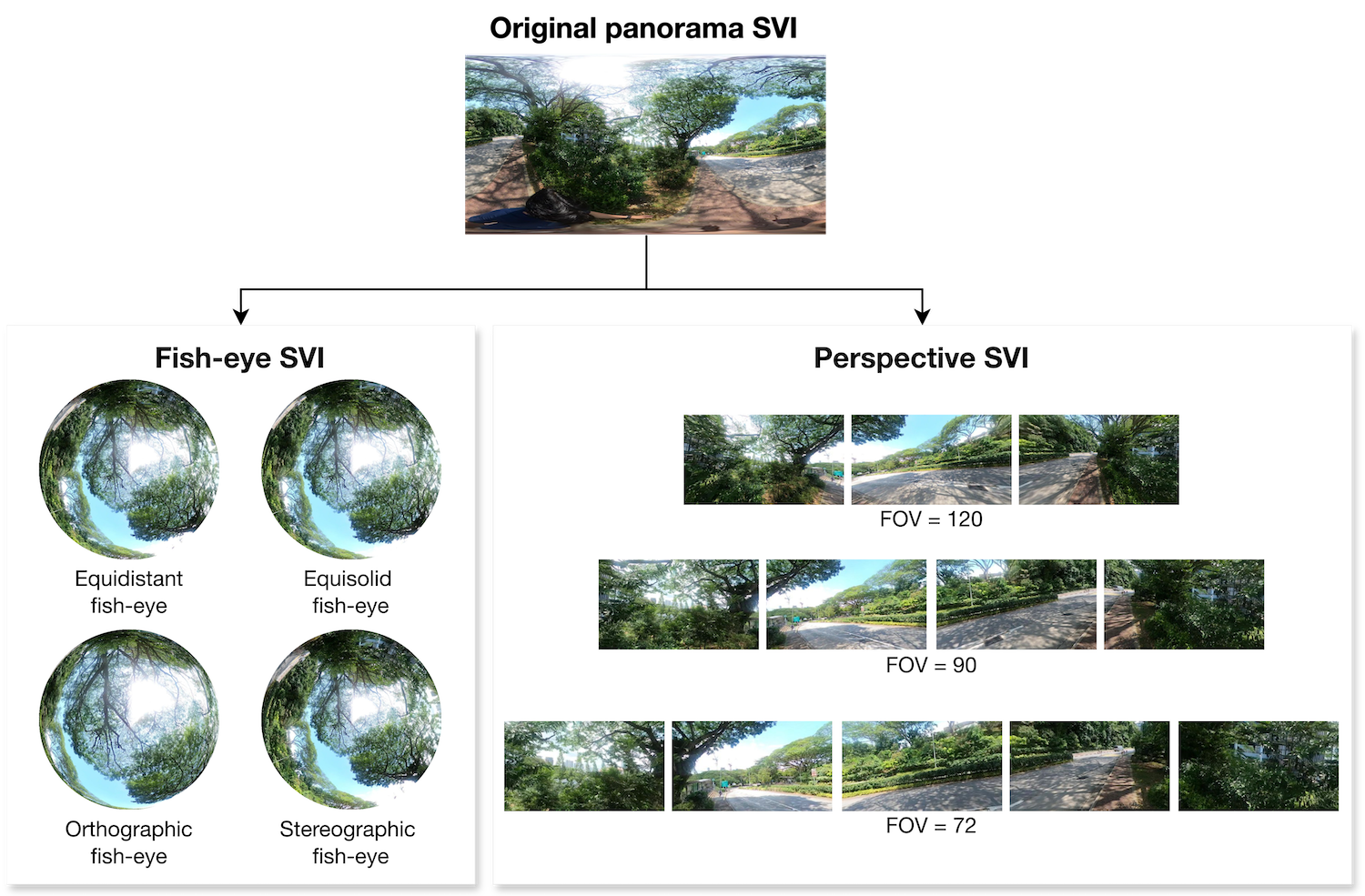}
    \caption{This figure illustrates different configurations of transformation from panorama SVI to fish-eye and perspective SVI. Source of imagery: Mapillary.}
    \label{fig:framework:fisheye_perspective}
\end{figure}

Transformation of SVI is also important because different image projections and data formats are required for different purposes ~\citep{2023_jag_svi_sensitivity}. 
Our sub-package for image transformation offers options to transform panoramic imagery into two further view types: `perspective' and `fisheye'. 
An overview and definitions of different image projections are provided in \autoref{fig:framework:fisheye_perspective}.

The left side of \autoref{fig:framework:fisheye_perspective} illustrates the four fisheye projection methods included in this sub-package: `orthographic', `equisolid angle', `equidistant', and `stereographic'. 
These projection methods have been employed across diverse research fields \citep{Schneider2009-er}. 
For example, the orthographic projection has been instrumental in calculating the sky view factor for urban microclimate studies \citep{Miao2020-yy, Carrasco-Hernandez2015-rp}, as the ratio of sky pixels in an upper fisheye photo with orthographic projection directly correlates to the diffuse solar irradiance absorbed by a horizontal surface \citep{Ivanova2019-cl}. 
This method also facilitates long-wave radiation transfer calculations between surfaces and surrounding objects, as the ratio of sky pixels to a surface's normal vector corresponds to the surface radiation \citep{Watanabe2013-ki}. 
For studies examining visual perceptions such as openness and oppressiveness \citep{Asgarzadeh2012-wm, Miyake2019-st}, the equisolid angle projection proves valuable because its pixel ratio directly corresponds to solid angles from a viewpoint. 
The equidistant projection finds its primary application in celestial position detection, as the distance between the center point and any other point in the fisheye photo directly corresponds to the angle in three-dimensional space \citep{Garcia-Gil2019-ix}.

The right side of \autoref{fig:framework:fisheye_perspective} displays three configurations for perspective SVI with fields of views (FOVs) at 72, 90, and 120 degrees. 
A perspective view is created by cropping an area with a specific direction and angle of view, which can be set as input arguments.
This view type simulates human vision when the angle of view accounts for the real ranges of human sight, making it suitable for both objective measurements of the environment and studies of subjective perception \citep{Kruse2021-rp, Wang2019-zj, Zhou2019-gq, Meng2020-fr}. 
The FOV setting significantly affects the resulting image: a lower FOV (e.g., 72 degrees in \autoref{fig:framework:fisheye_perspective}) creates less distortion but limits the captured area, whereas a higher FOV (e.g., 120 degrees) captures more area but introduces more distortion.

These diverse projection options enable researchers to select methods that best suit their specific research requirements and fields of study. 
The flexibility based on the parametrization of key settings contributes to meeting various analytical demands across different disciplines while ensuring reproducibility.

Our transformation sub-package also provides point cloud generation capabilities based on depth estimation in the computer vision sub-package, thus enabling advanced spatial analysis of street-level imagery. 
These techniques have evolved rapidly, particularly driven by autonomous vehicle applications \citep{li_deep_2020}, and can be utilized for various urban modeling applications \citep{biljecki_applications_2015}.

\begin{figure}[tbp]
    \includegraphics[width=\textwidth]{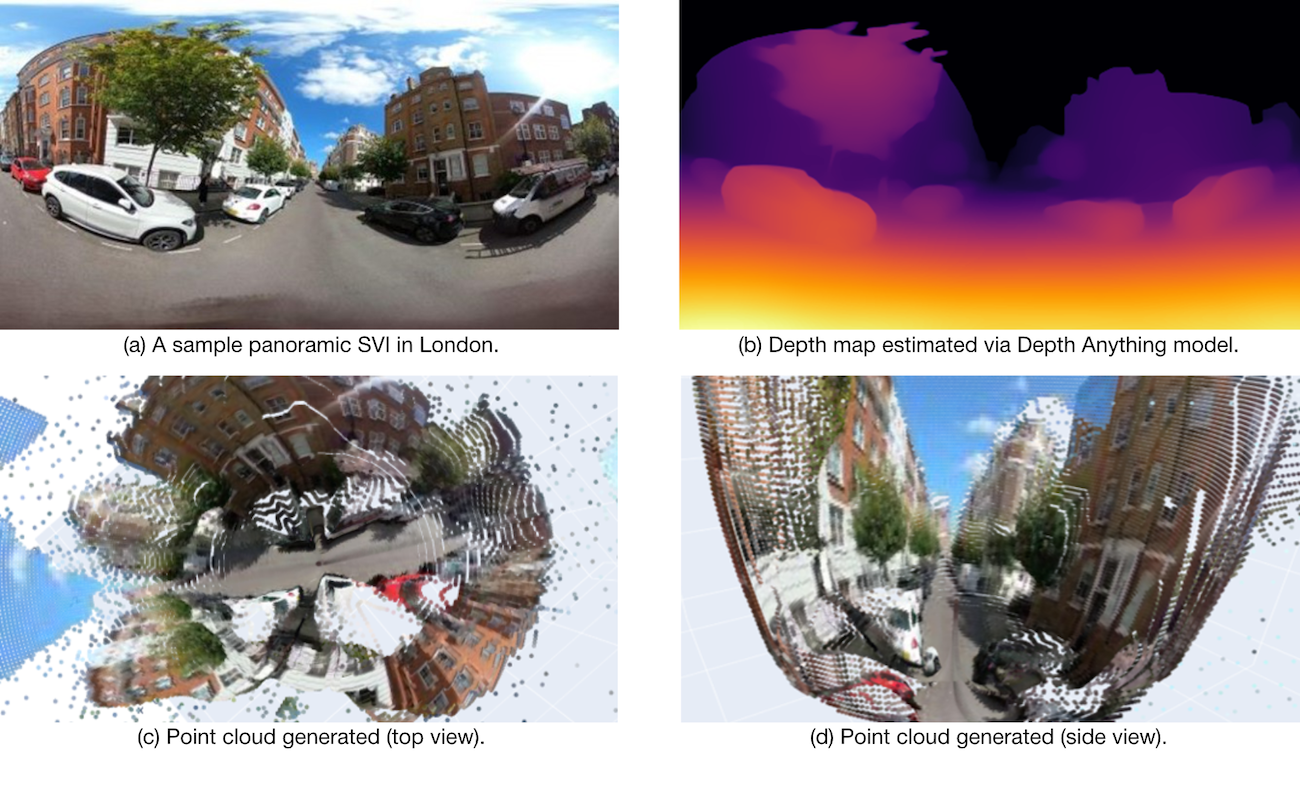}
    \caption{Example of depth estimation for a Mapillary SVI using the Depth Anything model and the resulting point cloud generation. Source of imagery: Mapillary.}
    \label{fig:framework:depth}
\end{figure}

For point cloud generation, we implemented the methodology of \citet{Cavallo20153DCR}, which combines depth estimates with RGB color information. 
The process projects pixels into spherical coordinates with depth-determined radii and then converts them to Cartesian coordinates for three-dimensional positioning. 
As demonstrated in \autoref{fig:framework:depth}, this workflow transforms panoramic SVI into detailed point clouds, enabling researchers to conduct advanced spatial analyses such as sidewalk width measurement and tree volume estimation \citep{tu_nerf2points_2024, orozcocarpio_3d_2024}. 
The visualization shows both top and side views, revealing the scene's outline, direction, and relative distances between street objects.

\subsection{Computer vision}\label{sc:cv}
\begin{figure}[tbp]
    \includegraphics[width=\textwidth]{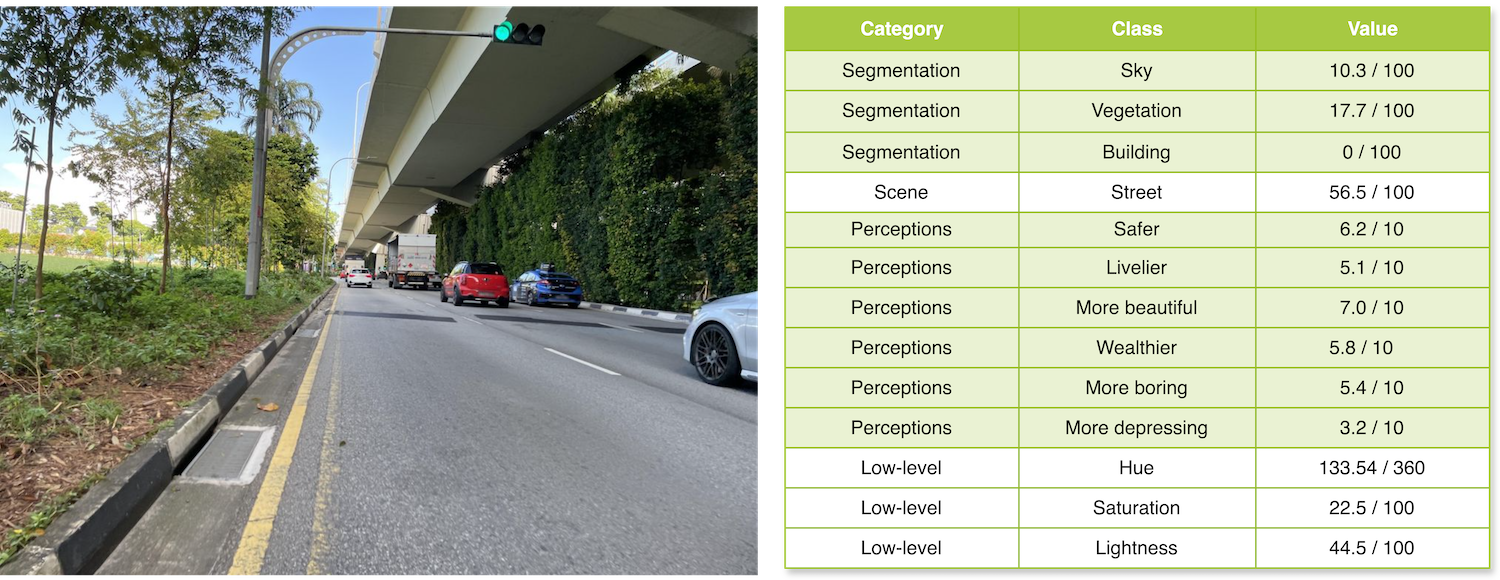}
    \caption{An example of scores and indicators on the right computed for the image on the left. Source of the image: Mapillary.}
    \label{fig:framework:example}
\end{figure}

Computer vision techniques have evolved into being indispensable for SVI data analysis. 
Our framework facilitates the use of well-established computer vision models by providing them with unified input and output formats. 
These models include object detection, scene segmentation, and feature extraction algorithms (see \autoref{framework:tab:cv}). \autoref{fig:framework:example} demonstrates sample outputs from these models, and their practical applications are further discussed in Section~\ref{sec:case_study}.

\begin{table}[ht]
\centering
\caption{Summary of computer vision models integrated in ZenSVI.}
\tiny
\begin{tabular}{p{0.2\textwidth} p{0.2\textwidth} p{0.4\textwidth}}
\toprule
\textbf{Model Type} & \textbf{Model Purpose} & \textbf{Description} \\
\midrule
Semantic/Panoptic Segmentation & Semantic Information Quantification & Mask2Former model pre-trained on Cityscapes and Mapillary Vistas datasets. Achieves mean IoU of 84.5\% (semantic) and 64.7\% (panoptic) for semantic segmentation and 66.6\% and 45.5\% for panoptic segmentation \citep{cheng_maskedattention_2022}. \\
\midrule
Image Classification & Perception Classification & Place Pulse 2.0 model trained using ResNet50 architecture. Measures human perception of streetscapes across six dimensions. Achieves an average top-2 accuracy of 66.03\% \citep{dubey2016deep}. \\

Image Classification & Scene Classification & Places365 model trained using ResNet architecture. Classifies 365 scene categories and 102 scene attributes. Achieves 85.08\% precision for the top-5 scene categories \citep{zhou_places_2017}. \\

Image Classification & Weather Classification & Model trained on \citet{hou2024global} dataset. Classifies weather in SVI as clear, cloudy, foggy, rainy, or snowy \citep{hou2024global}. \\

Image Classification & Glare Detection & Model trained on \citet{hou2024global} dataset. Detects glare in SVI (True or False) \citep{hou2024global}. \\

Image Classification & Time of Day Classification & Model trained on \citet{hou2024global} dataset. Classifies time of day in SVI as day, dusk/dawn, or night \citep{hou2024global}. \\

Image Classification & Panorama Detection & Model trained on \citet{hou2024global} dataset. Detects if SVI is a panorama (True or False) \citep{hou2024global}. \\

Image Classification & Platform Classification & Model trained on \citet{hou2024global} dataset. Classifies platform of SVI as cycling surface, driving surface, fields, railway, tunnel, or walking surface \citep{hou2024global}. \\

Image Classification & Quality Classification & Model trained on \citet{hou2024global} dataset. Classifies quality of SVI as good, slightly poor, or very poor \citep{hou2024global}. \\

Image Classification & Reflection Detection & Model trained on \citet{hou2024global} dataset. Detects reflection in SVI (True or False) \citep{hou2024global}. \\

Image Classification & View Direction Classification & Model trained on \citet{hou2024global} dataset. Classifies view direction of SVI as front/back or side \citep{hou2024global}. \\
\midrule
Low-level Features Extraction & Edge Detection & Canny, SobelX, SobelY, and Laplacian techniques included. Detects edges of objects in images \citep{canny_computational_1986, sobel_3x3_1968, marr_theory_1997}. \\

Low-level Features Extraction & Blob Detection & Identifies regions in the image that differ in properties like brightness or color compared to surrounding regions \citep{lindeberg_feature_1998}. \\

Low-level Features Extraction & Blur Detection & Determines the blurriness of an image for quality control \citep{gonzalez_digital_2006}. \\

Low-level Features Extraction & Color Space Analysis & Analyzes color characteristics in hue, saturation, and lightness. \\
\midrule
Image Embedding & Feature Extraction & Converts images into vector representation using pre-trained models like ResNet, VGG, DenseNet, AlexNet, and EfficientNet for various tasks such as image retrieval, clustering, and classification \citep{he2016deep, simonyan2014very, huang2017densely, krizhevsky2012imagenet, tan2019efficientnet}. \\
\bottomrule
\end{tabular}
\label{framework:tab:cv}
\end{table}

Our package provides both semantic and panoptic segmentation capabilities. While semantic segmentation classifies pixels into categories, panoptic segmentation additionally identifies unique object instances, enabling tasks like counting individual trees or cars --- a crucial feature for scalable urban analysis \citep{S_nchez_2024}. 
We use the Mask2Former model \citep{cheng_maskedattention_2022}, pre-trained on two key urban datasets. 
The model achieves 84.5\% mIoU for semantic and 66.6\% for panoptic segmentation on Cityscapes, 64.7\% mIoU for semantic and 45.5\% for panoptic segmentation on Mapillary Vistas.

\begin{figure}[tbp]
   \includegraphics[width=\textwidth]{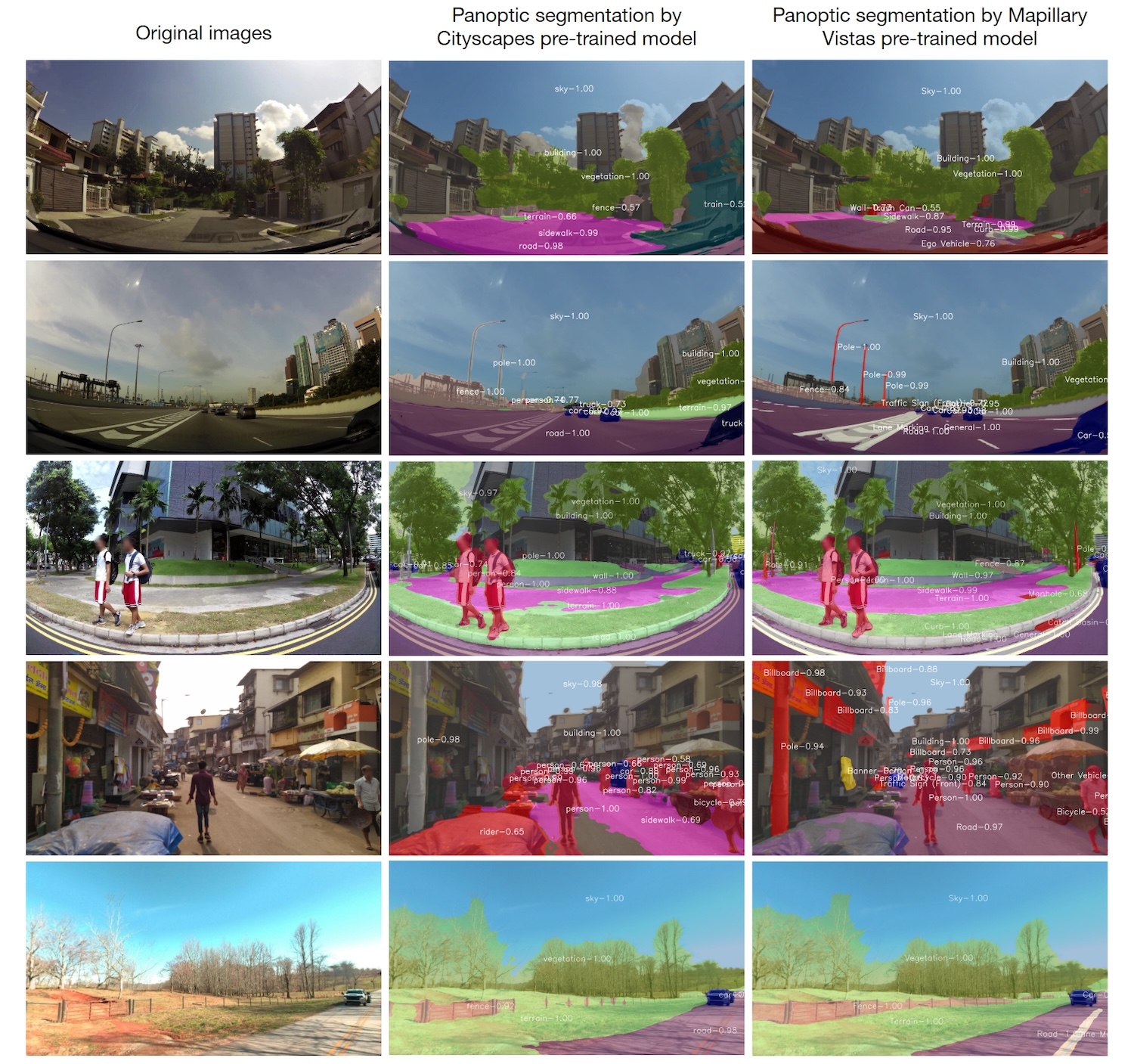}
   \caption{Panoptic segmentation results of five Mapillary images using Cityscapes and Mapillary Vistas pre-trained models. The dual-model approach enhances adaptability to diverse urban contexts: Mapillary Vistas offers specialized classes (e.g., bike lanes) for detailed analysis, while Cityscapes provides broader classes for general urban assessment.}
   \label{fig:framework:segmentation}
\end{figure}

As demonstrated in \autoref{fig:framework:segmentation}, the segmentation results clearly identify and color-code various street features including buildings, roads, sidewalks, and vegetation. 
The package outputs pixel ratios for all labels to a CSV/JSON file, facilitating quantitative analysis \citep{ibrahim2020understanding}. 
This functionality enables measurement of street elements across different locations and periods \citep{ye_visual_2019, ramirez_measuring_2021, ma_measuring_2021}, supporting applications such as green view index calculation --- a measure of visible greenery at eye-level \citep{li_assessing_2015, yang_can_2009}.

Our package provides several pre-trained image classification models, with particular emphasis on street-level perception analysis. 
A key model is trained on the Place Pulse 2.0 dataset \citep{dubey2016deep}, which comprises 110,988 images from 56 cities across 28 countries. 
This dataset captures the human perception of streetscapes along six dimensions: depressing, boring, beautiful, safe, lively, and wealthy, based on volunteer-provided labels. 
Following the methodology of \citet{kang2023assessing}, we implemented a ResNet50 architecture where perceptual scores are categorized into five classes. 
The models achieved an average top-2 accuracy of 66.03\% across all six perceptual dimensions. 
When processing input images, the package generates a five-dimensional probability vector for each perceptual category, which is then converted into final dimensional scores ranging from 0 to 10.
An existing model used in Global Streetscapes \citep{hou2024global} was also added for benchmarking purposes.
This model has a Vision Transformer (ViT) architecture and also generates perception scores, ranging from 0 to 10, with an average accuracy ranging from 61.6\% to 77.1\% for all six perceptual dimensions.

\begin{figure}[tbp]
    \includegraphics[width=\textwidth]{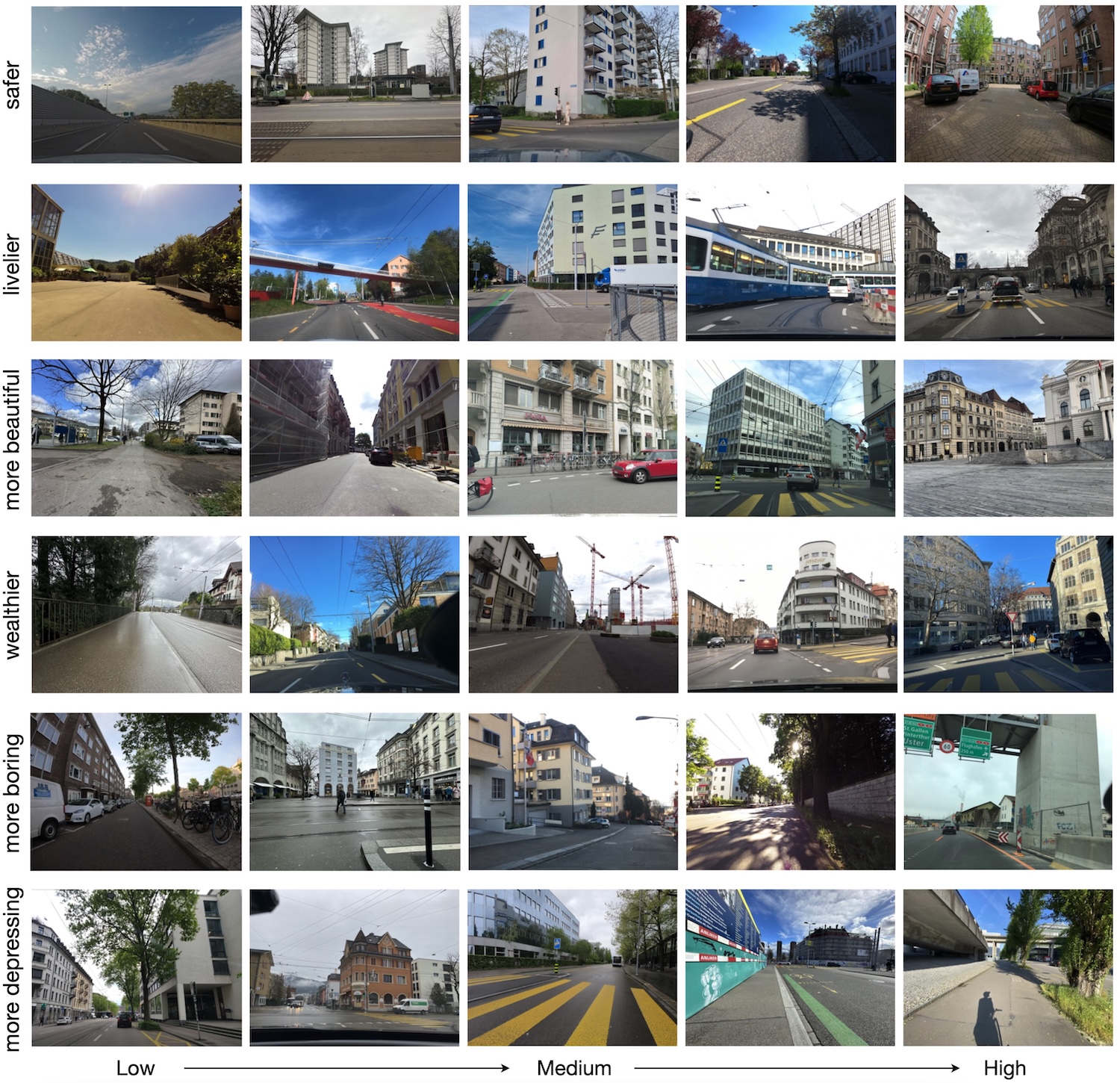}
    \caption{Image samples from Mapillary evaluated using human perception models, ranked from lower (left) to higher (right) human perception scores across six dimensions.}
    \label{fig:framework:perception}
\end{figure}

This perceptual analysis framework has found widespread applications, from urban change tracking \citep{liang_revealing_2023, ito_understanding_2024} to public housing evaluation \citep{wang_assessing_2024}. 
\autoref{fig:framework:perception} presents ranked image samples across the six perceptual dimensions and reveals that locations featuring diverse urban elements (e.g., buildings, vehicles, vegetation) typically score higher in safety and liveliness metrics. 
Conversely, construction sites and monotonous street designs tend to receive higher scores for boringness and depressiveness. 
This human-centric evaluation approach enables researchers to quantitatively assess the built environment's perceptual qualities and integrate these insights into broader urban studies.

ZenSVI incorporates a scene classification model pre-trained on the Places365 dataset by \citet{zhou_places_2017}, which achieves 85.08\% top-5 precision in scene category classification. 
This comprehensive dataset contains 365 distinct scene categories (such as residential neighborhoods, building facades, and field roads) and 102 scene attributes (including features like biking areas, asphalt surfaces, and grass coverage). 
The model's usefulness has been validated through numerous research applications, from environmental feature extraction \citep{luo_geotile2vec_2023, yang_public_2024} to targeted scene filtering for specific urban analyses \citep{chen_quantifying_2020, deoliveira_outdoorsent_2020, seresinhe_using_2017}.

\begin{figure}[tbp]
    \includegraphics[width=\textwidth]{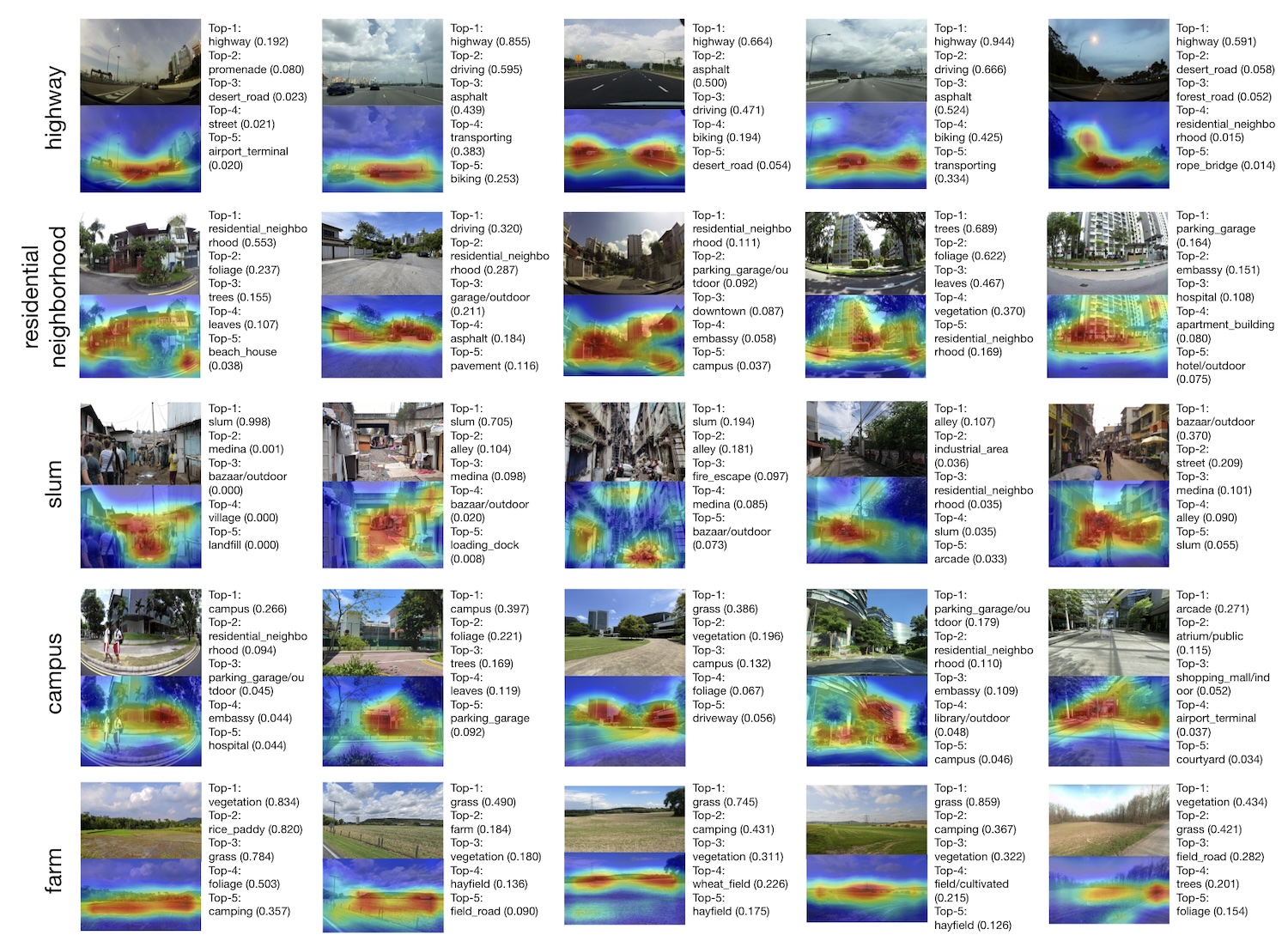}
    \caption{Places365 scene classification examples in five different settings: highway, residential neighborhood, slum, campus, and farm.}
    \label{fig:framework:places365}
\end{figure}

As demonstrated in \autoref{fig:framework:places365}, our implementation excels at analyzing SVI across diverse urban and rural settings, including highways, residential neighborhoods, slums, campuses, and farms. 
For each image, the model generates both probability distributions across scene categories and class activation mapping (CAM) heatmaps. 
These CAM heatmaps visually reveal which regions of the image most significantly contribute to the scene classification \citep{zhang_measuring_2018}. 
For example, the model focuses on road infrastructure and traffic patterns when classifying highways, while for residential neighborhoods, it attends to architectural features, landscaping elements, and the overall spatial organization of buildings and vegetation.
This functionality enables researchers to conduct detailed environmental analyses through automated scene recognition \citep{chen2020quantifying, yang_public_2024}.

In our computer vision sub-package, we also integrated image classification models developed by \citet{hou2024global}, trained on the manually labeled subset of the NUS Global Streetscapes dataset comprising more than 10,000 images from over 400 cities across all continents. 
These images were sourced from Mapillary and KartaView platforms. 
Our sub-package has eight distinct classification models that analyze the following image attributes:

\begin{itemize}
    \item Weather conditions (clear, cloudy, foggy, rainy, or snowy)
    \item Glare presence (True or False)
    \item Time of day (day, dusk/dawn, or night)
    \item Panorama status (True or False)
    \item Platform type (cycling surface, driving surface, fields, railway, tunnel, or walking surface)
    \item Image quality (good, slightly poor, or very poor)
    \item Reflection presence (True or False)
    \item View direction (front/back or side)
\end{itemize}

\begin{figure}[tbp]
    \includegraphics[width=\textwidth]{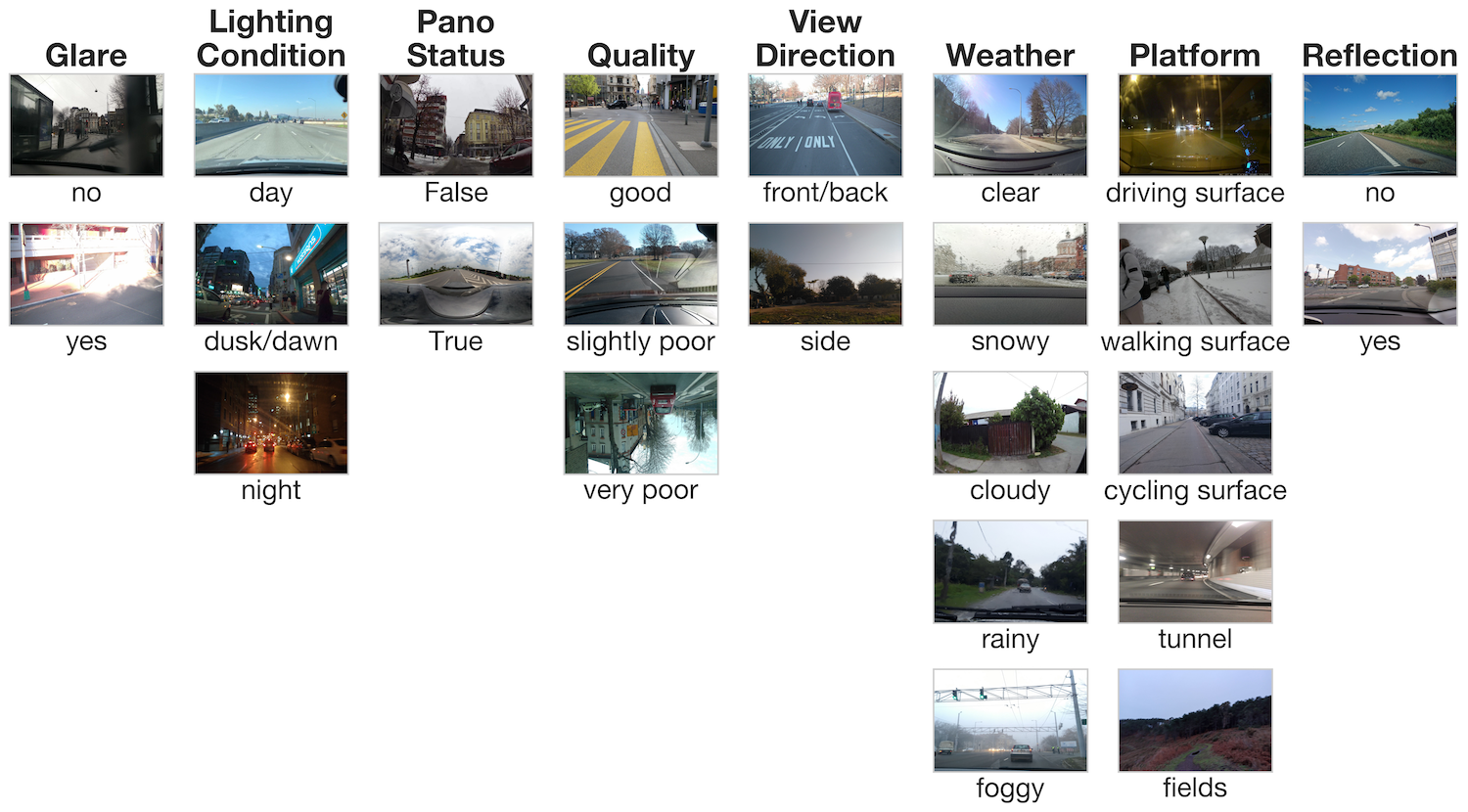}
    \caption{This figure shows examples of SVI classified into different conditions, under which they were captured. Columns from left to right show glare, lighting condition, pano status, quality, view direction, weather, platform, and reflection.}
    \label{fig:framework:gss}
\end{figure}

As illustrated in \autoref{fig:framework:gss}, these models enable comprehensive filtering of SVI based on specific research requirements. 
For instance, researchers studying nighttime pedestrian experiences can isolate high-quality images taken from walking surfaces during nighttime conditions \citep{hou2024global}. 
This systematic filtering capability helps mitigate potential biases in SVI analysis.

For object detection tasks, we integrate GroundingDINO in ZenSVI. This state-of-the-art model can detect arbitrary objects based on human inputs such as category names or referring expressions \citep{liu2023grounding}. Trained on several widely used object detection datasets --- including COCO, O365, and Open Images --- GroundingDINO has also been implemented in urban research, enabling large-scale analyses of built environment features from SVI \citep{liang2024evaluating, ramalingam2025building}. By specifying category names (e.g., `tree', `building', or multiple objects such as `person . car .'), this open-set detector can generate bounding boxes for the identified objects in images. 
This open-set capability enables the quantification of urban elements and offers flexibility for a wide range of applications.

\begin{figure}
    \centering
    \includegraphics[width=\linewidth]{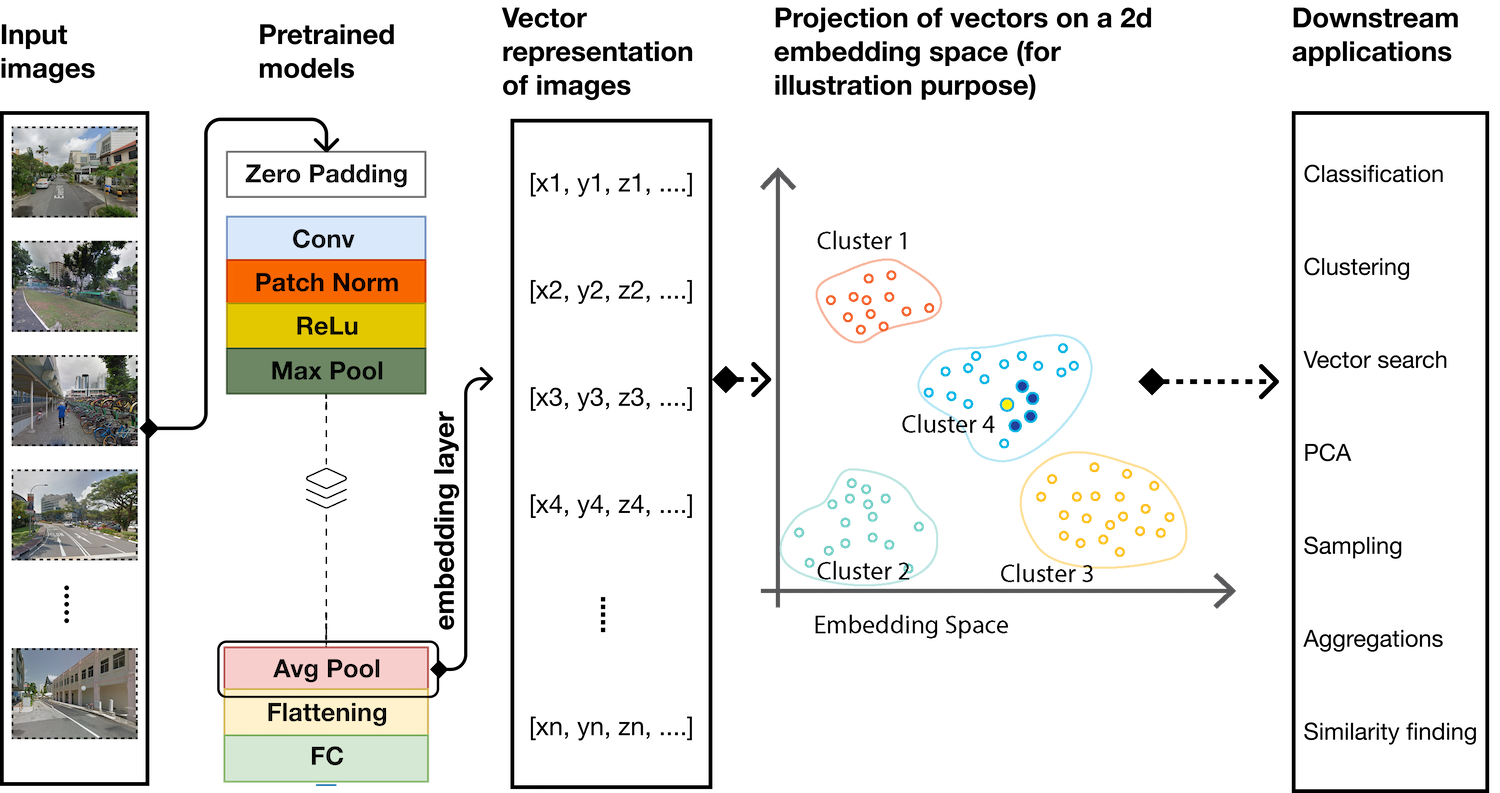}
    \caption{Extracting image embeddings from pre-trained Models. The image embeddings are extracted from the last layer of the pre-trained models, capturing the high-level features of the images. The embeddings are then used to calculate the similarity between images and identify clusters of images with similar features.}
    \label{fig:extracting_image_embeddings}
\end{figure}

\begin{figure}
    \centering
    \includegraphics[width=0.6\linewidth]{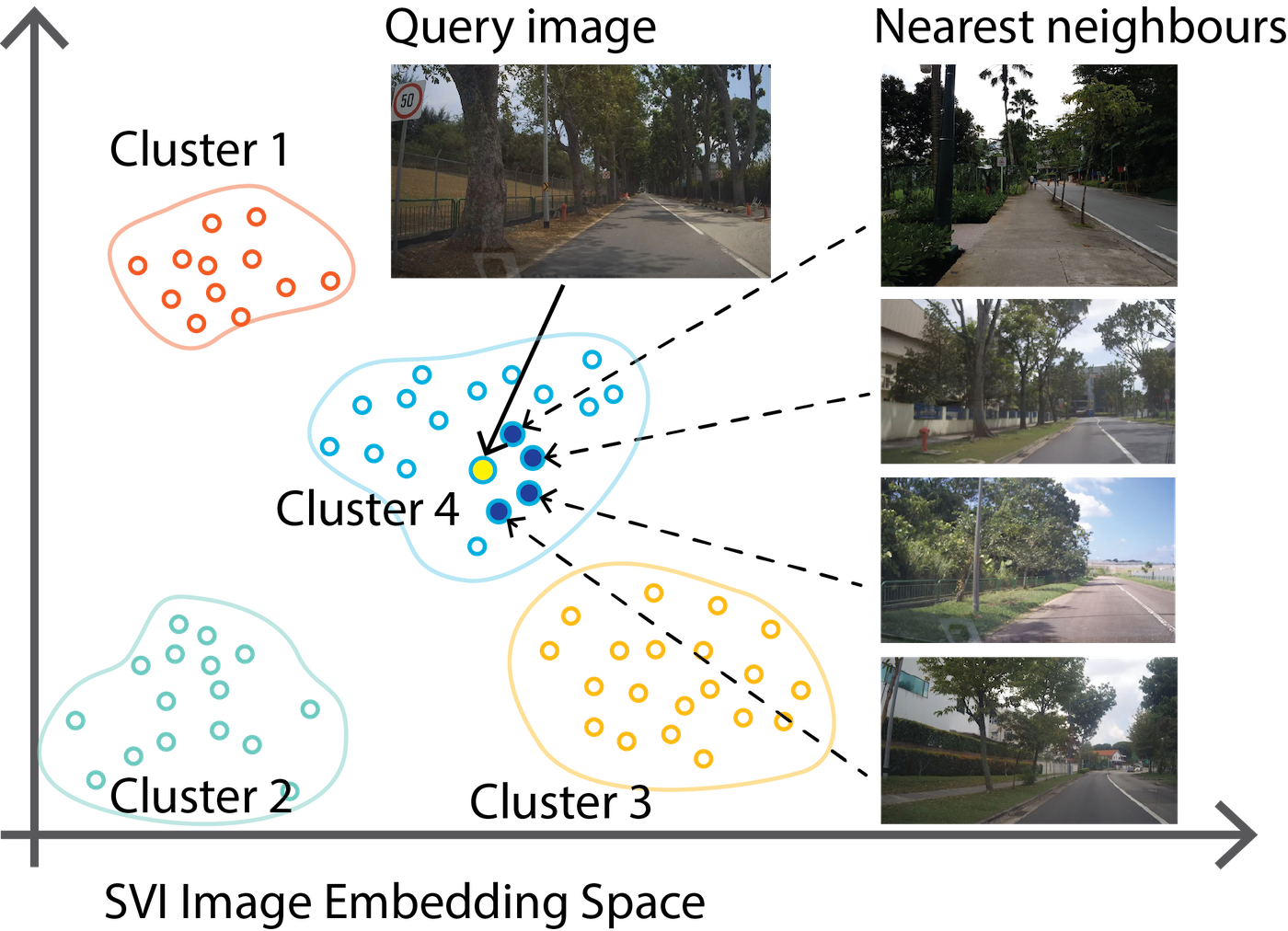}
    \caption{Illustration of image embeddings in embedding space. Each image can be represented as a multi-dimensional vector. Each vector can be projected in the embedding space, the distance between each vector in the space indicates the similarity between features of the images.}
    \label{fig:embeddings}
\end{figure}

Our sub-package also offers image embedding capabilities, leveraging various pre-trained models including ResNet, VGG, DenseNet, AlexNet, and EfficientNet \citep{he2016deep, simonyan2014very, huang2017densely, krizhevsky2012imagenet, tan2019efficientnet}. 
These models transform SVI into multi-dimensional vector representations that capture their essential features. 
The embedding process extracts high-level features from the models' final layers. 
For instance, ResNet generates a 2048-dimensional vector from its average pooling layer, as illustrated in \autoref{fig:extracting_image_embeddings}. 
To quantify image similarities, we implemented cosine distance metrics, defined as:

\begin{equation}
    \text{Cosine Distance} = 1 - \frac{\text{A} \cdot \text{B}}{\|\text{A}\| \times \|\text{B}\|}
\end{equation}

where A and B represent the embedding vectors being compared. 
This metric ranges from 0 (identical) to 1 (completely different), enabling efficient image comparison and clustering. 
As demonstrated in \autoref{fig:embeddings}, these embeddings can be projected into a two-dimensional space for visualization, where proximity between points indicates feature similarity. 
This capability facilitates various applications, including image retrieval \citep{jush_medical_2023}, clustering \citep{abrar_effectiveness_2023}, and pattern identification \citep{ypsilantis_universal_2023} in SVI. 
The technique proves particularly valuable for efficiently sampling representative images while eliminating redundant ones in downstream applications.

The sub-package also supports the following two types of depth estimation:
\begin{itemize}
    \item Relative depth estimation: Returns normalized distances (0-255) for each pixel using the Dense Prediction Transformer \citep{ranftl_vision_2021}.
    \item Absolute depth estimation: Provides distances in meters using the Depth Anything model \citep{yang_depth_2024}.
\end{itemize}

Lastly, our computer vision sub-package includes comprehensive capabilities for extracting low-level image features, which are crucial for applications such as human perception prediction \citep{guan_urban_2021a, dubey2016deep}. 
The package implements several key feature detection techniques:

\textbf{Edge Detection:}
We provide four distinct edge detection methods:
\begin{itemize}
    \item Canny detection for precise edge localization \citep{canny_computational_1986}.
    \item SobelX and SobelY detection for directional edge identification (horizontal and vertical respectively) \citep{sobel_3x3_1968}.
    \item Laplacian detection for emphasizing rapid intensity changes and identifying blobs or junctions \citep{marr_theory_1997}. The Laplacian operator is defined as:
    \begin{equation}
        \nabla^2 f = \frac{\partial^2 f}{\partial x^2} + \frac{\partial^2 f}{\partial y^2}
    \end{equation}
    which is implemented using the following discrete kernel:
    \begin{equation}
        K = \begin{bmatrix} 
        0 & 1 & 0 \\
        1 & -4 & 1 \\
        0 & 1 & 0
        \end{bmatrix}
    \end{equation}
\end{itemize}

\textbf{Additional Feature Analysis:}
\begin{itemize}
    \item Blob detection for identifying regions with distinct properties (brightness, color) relative to their surroundings \citep{lindeberg_feature_1998}.
    \item Blur detection for image quality assessment \citep{gonzalez_digital_2006}.
    \item Color space analysis in HSL (Hue, Saturation, Lightness) format, providing a more intuitive representation of color characteristics compared to RGB space.
\end{itemize}

\subsection{Visualization}\label{sc:visualization}
The visualization sub-package, built with Matplotlib \citep{Hunter:2007}, provides four high-level plotting functions to streamline the presentation of SVI analysis for quick initial data exploration. 
ZenSVI supports histogram generation for discrete data visualization, such as object detection counts, and kernel density estimation plots for continuous variables, such as semantic segmentation ratios. 
Users can create image grids to display SVI samples, either randomly selected or sorted by specified variables.

For spatial visualization, the package implements three map formats: point maps showing individual SVI locations, line maps using OpenStreetMap street segments \citep{boeing_osmnx_2017}, and hexagonal maps using Uber's H3 grid \citep{uber_h3_2018}. 
In both line and hexagonal maps, SVI values are aggregated to the nearest features, defaulting to point counts when no specific variable is specified.

\begin{figure}[tbp]
    \centering
    \begin{subfigure}{.45\textwidth}
        \centering
        \includegraphics[width=\textwidth]{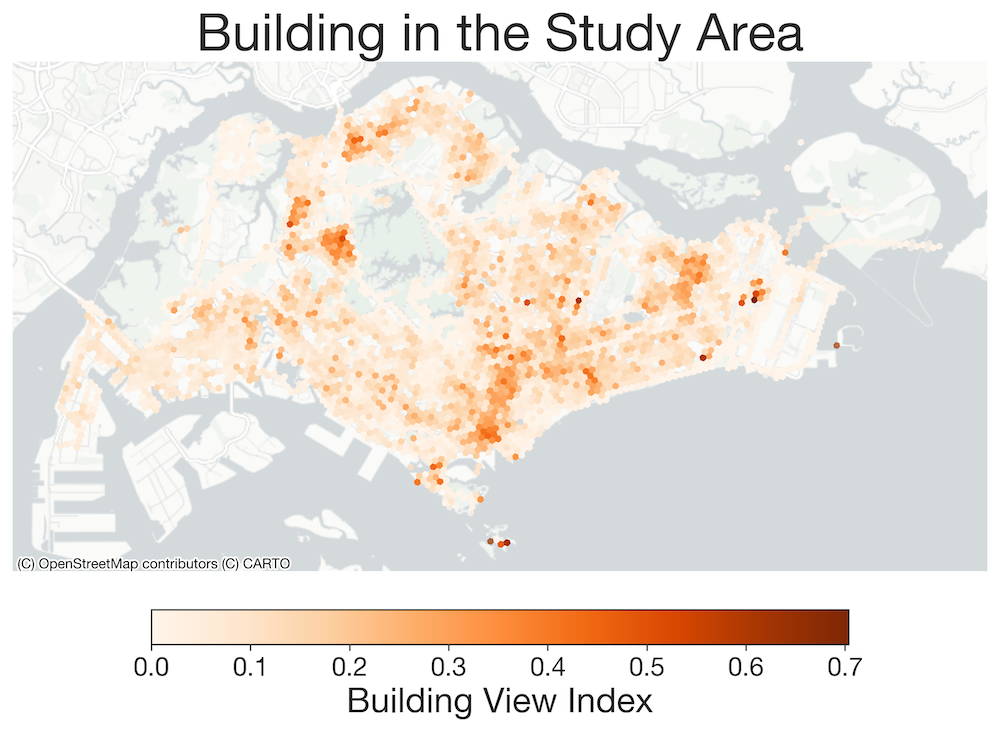}
        \caption{A map of building view index in Singapore.}
    \end{subfigure}
    \hfill
    \begin{subfigure}{.45\textwidth}
        \centering
        \includegraphics[width=\textwidth]{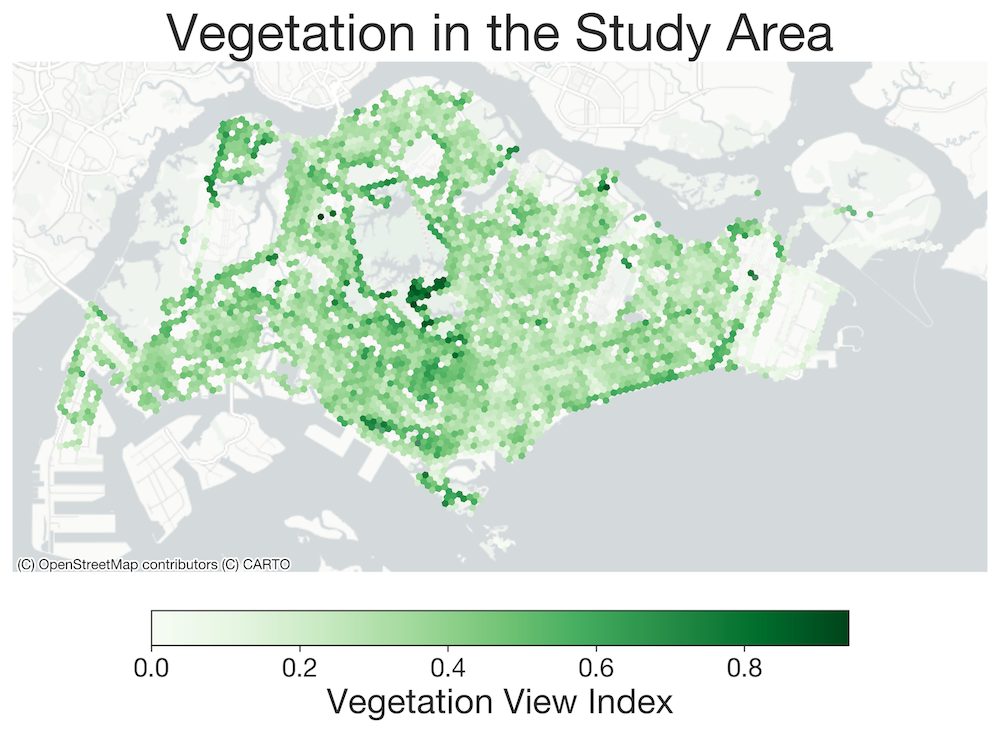}
        \caption{A map of vegetation view index in Singapore.}
    \end{subfigure}
    \begin{subfigure}{.45\textwidth}
        \centering
        \includegraphics[width=\textwidth]{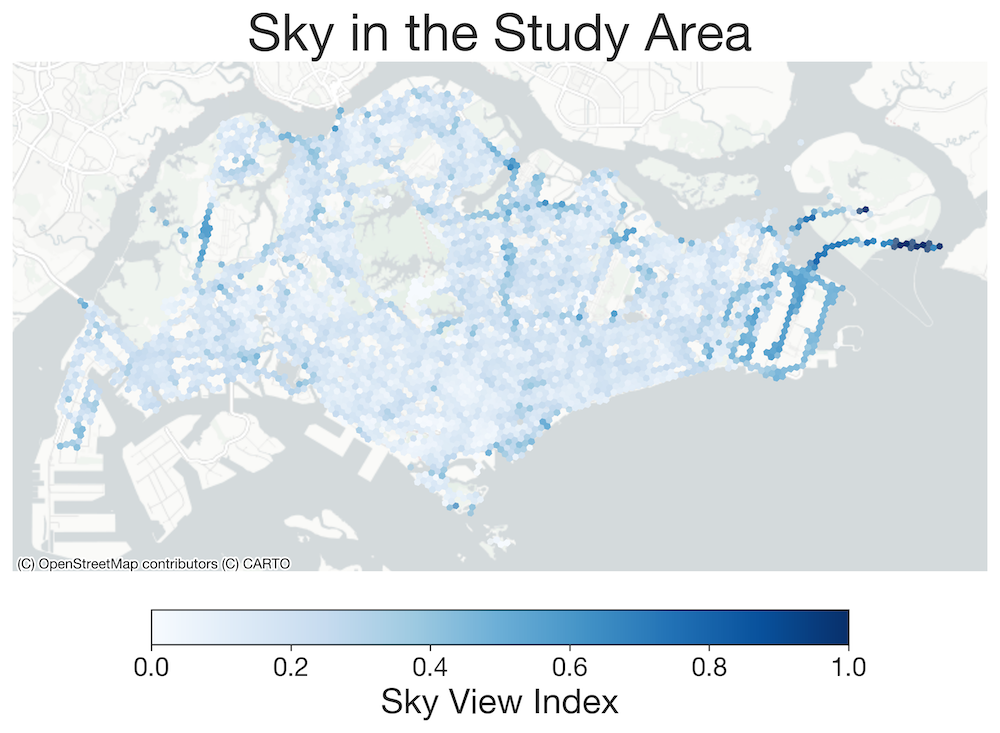}
        \caption{A map of sky view index in Singapore}
    \end{subfigure}
    \hfill
    \begin{subfigure}{.45\textwidth}
        \centering
        \includegraphics[width=\textwidth]{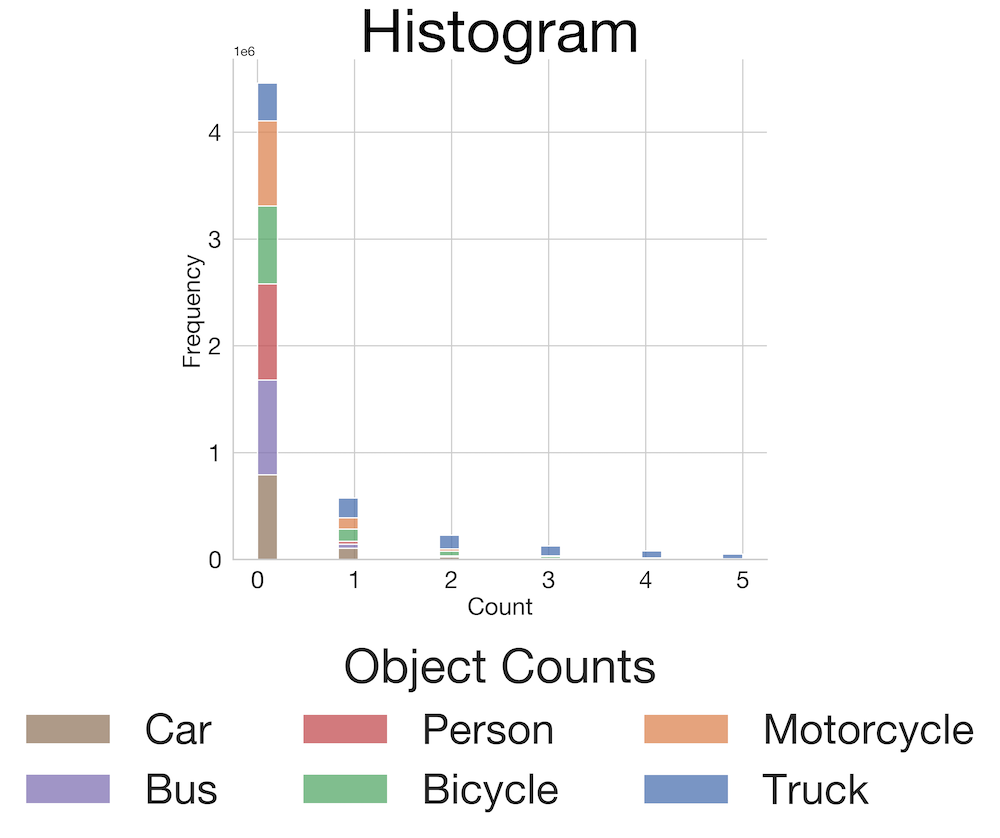}
        \caption{Frequency distribution of detected urban objects (vehicles, people, and bicycles) in Singapore street-level imagery. The histogram shows the count distribution for different object categories, with most observations having 0-1 instances per image.}
    \end{subfigure}
    \begin{subfigure}{.45\textwidth}
        \centering
        \includegraphics[width=\textwidth]{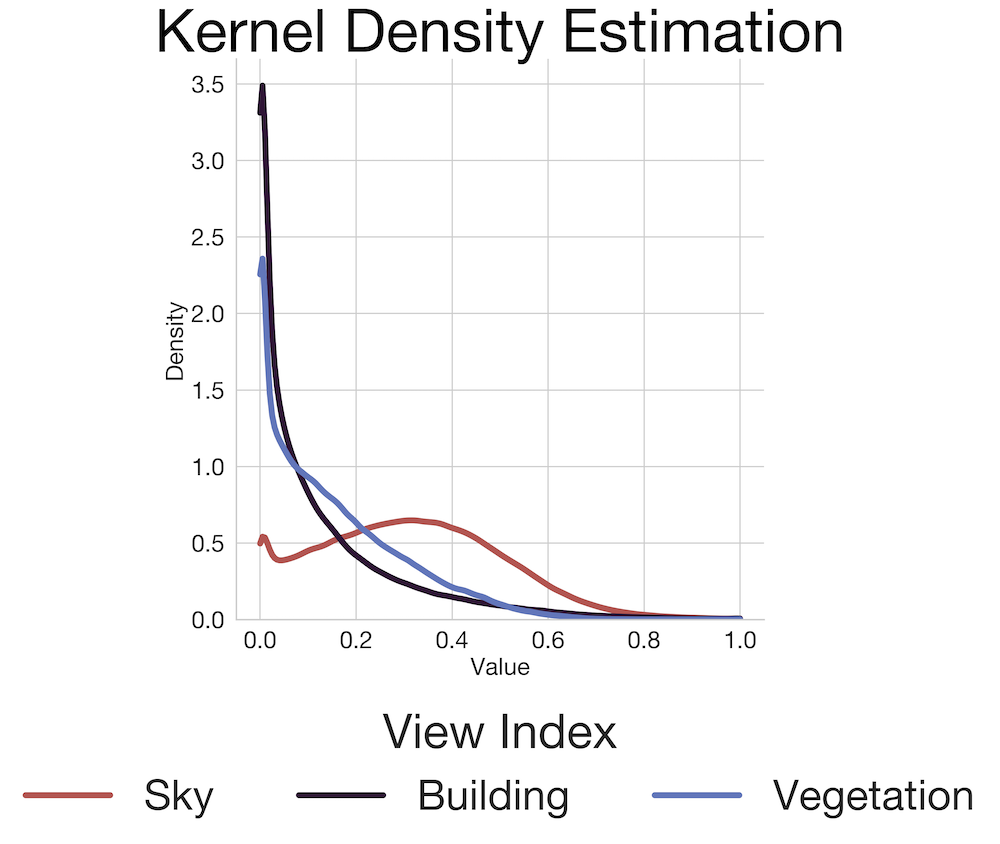}
        \caption{Probability density distribution of view indices across Singapore. The plot shows the relative frequency of different index values [0, 1] for buildings, vegetation, and sky visibility, highlighting the varying urban morphology patterns. Building views show high density at lower values, while sky views demonstrate a bimodal distribution.}
    \end{subfigure}
    \caption{Examples of how our visualization sub-package can be used to plot maps, histograms, and kernel density estimation plots.}
    \label{fig:framework:visualization}
\end{figure}

These visualization tools effectively communicate spatial patterns in urban environments. 
For example, \autoref{fig:framework:visualization} reveals higher building view indices in the central business district and residential areas, elevated vegetation indices in central forested regions, and increased sky view indices along coastal and riverfront areas. 
The distribution plots highlight the varying patterns of transportation modes and view indices across the study area. 
All the functions return figure and axis instances from the Matplotlib package so that the users can further modify the plots flexibly. 

\section{Examples and case study} \label{sec:case_study}
In this section, we conducted a comprehensive case study to showcase examples of how our package can be used to help researchers in one of the most typical studies by exhibiting each functionality in our package. 
We used Singapore as the study site and aim to investigate the daytime urban characteristics by conducting \textit{k}-means clustering based on the features extracted from SVI obtained from Mapillary. 
This analysis was chosen because this type of analysis with SVI has been implemented for various purposes by researchers \citep{fan_nighttime_2024, gao_integrating_2024, zhang_efficient_2023, liang_revealing_2023, wang_investigating_2024}, and it also includes data engineering processes such as filtering unsuitable imagery. 
This process was conducted as follows:
\begin{enumerate}
    \item Download SVI from Mapillary.
    \item Apply computer vision models to extract features.
    \item Conduct quality assessment and remove unsuitable SVI.
    \item Run \textit{k}-means clustering analysis.
\end{enumerate}

\subsection{Download and quality control}
We downloaded SVI from Mapillary by simply inputting `Singapore' to the \texttt{download\_svi} function, which was then passed to OSMnx to retrieve the boundary of Singapore. 
This yielded 932,596 images as of June 2024. 
Then, we applied the following computer vision models to extract features used for the subsequent cleaning process and analysis:
\begin{itemize}
    \item Image classification model trained on Places365.
    \item Image classification model trained on PlacePulse.
    \item Image classification model trained on NUS Global Streetscape dataset (e.g., image quality, platform, etc).
    \item Semantic segmentation trained on Mapillary Vistas.
    \item Low-level image processing (e.g., blur detection, etc).
\end{itemize}

In addition to these extracted features, we also used metadata from Mapillary for the cleaning process, such as the speed and timestamp of the device when the image was taken. 
After extracting the necessary features, the dataset was cleaned using a two-stage filtering approach to ensure images were both of high quality and suitable for our specific use case. 
For this case study focusing on characterizing the urban environment during daytime, we applied the following criteria:

\begin{enumerate}
   \item \textbf{Image Quality Requirements}
   \begin{enumerate}[label=\alph*)]
       \item \textbf{Overall Quality}: Images labeled as ``very poor'' by the image quality classification model were excluded.
       \item \textbf{Sharpness}: Blurry images with Laplacian variance over 100 were removed.
   \end{enumerate}
   
   \item \textbf{Image Properties for Use Case Suitability}
   \begin{enumerate}[label=\alph*)]
       \item \textbf{Visual Complexity}: A minimum visual complexity score of 1.0 was required, calculated using the Shannon Index:
       \begin{equation}
       H = -\sum_{i=1}^{n} p_i \ln(p_i)
       \end{equation}
       where $p_i$ is the normalized ratio of pixels for segment $i$, and $n$ is the number of different segments in the image. The normalization ensures $\sum_{i=1}^{n} p_i = 1$. This measure, originally from information theory, is now used to clean up SVI datasets in urban studies \citep{yap_urbanity_2023}.
       \item \textbf{Speed Criterion}: The device speed was required to be $\leq 200$ km/h when available in the metadata. We applied this step to remove outliers, such as images taken from an airplane, which may also indicate positional inaccuracy due to GPS lag at high velocities.
       \item \textbf{Lighting Conditions}: Images were filtered to exclude nighttime shots based on the NUS Global Streetscape's lighting model and the metadata.
       \item \textbf{View Direction}: Only images labeled as ``front/back'' viewing by the NUS Global Streetscape's image quality classification model were retained.
       \item \textbf{Platform}: Based on the NUS Global Streetscape's platform classification model, we only kept images labeled as being recorded from the driving surface.
   \end{enumerate}
\end{enumerate}

\begin{figure}
    \centering
    \includegraphics[width=\linewidth]{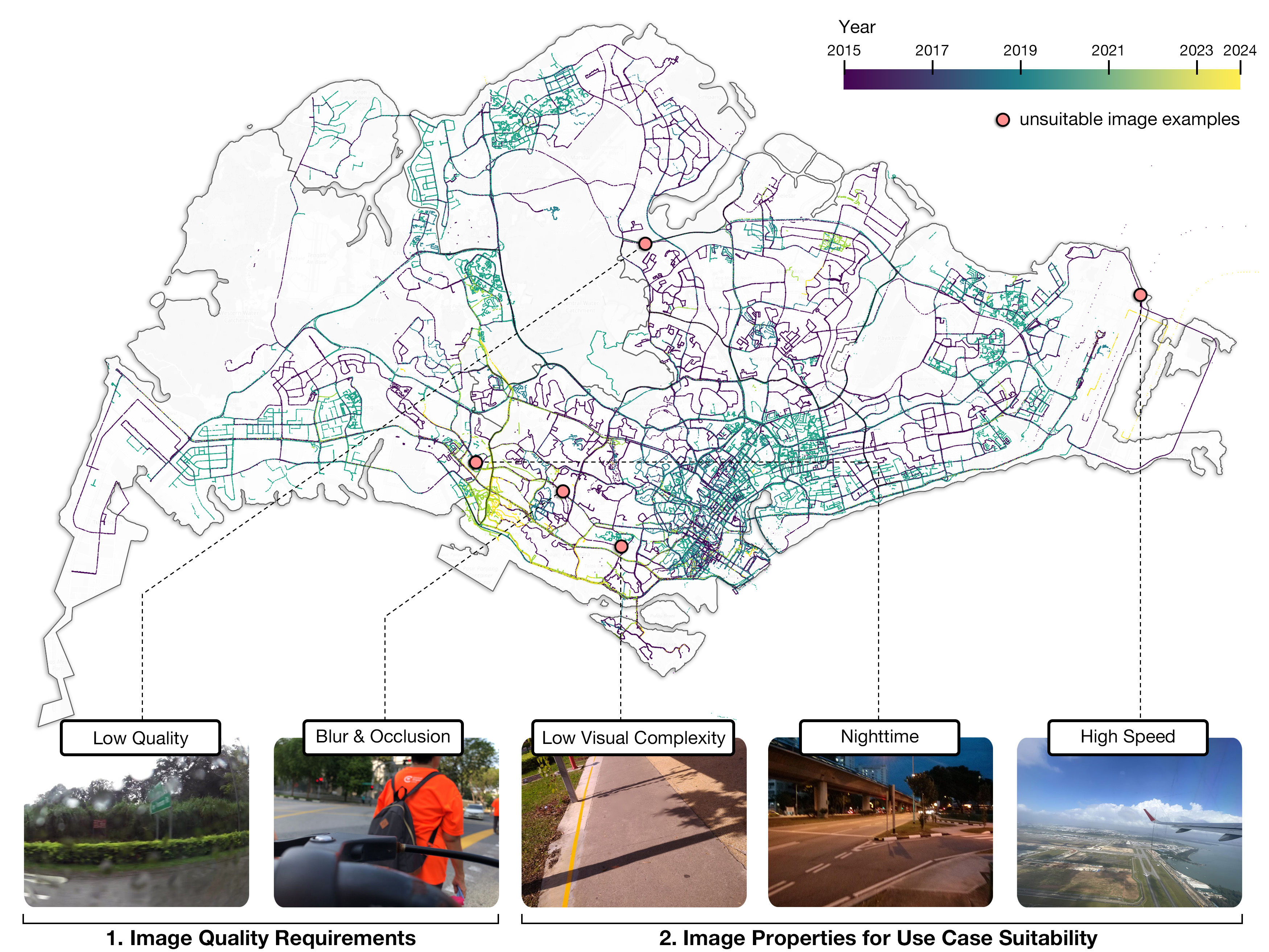}
    \caption{A map of suitable SVI points in Singapore downloaded from Mapillary. The color represents the year of being recorded. This map also shows some examples of unsuitable images removed from the dataset, including nighttime images, images with blur, low visual complexity, and low quality, and images taken at extremely high speeds.}
    \label{fig:case_study:invalid_map}
\end{figure}

By following these criteria, we targeted the daytime SVI with good quality and sufficient visual information.
For each criterion, missing values (\texttt{NA}) were treated as suitable to avoid excessive data loss. 
The final suitability flag (\texttt{is\_suitable}) was computed as the logical conjunction of all individual criteria, meaning an image needed to pass all applicable filters to be considered suitable. 
After the cleaning process, 270,864 images ---  about 30\% of the initial dataset --- remained. 
As displayed in \autoref{fig:case_study:invalid_map}, our suitable SVI points cover most of Singapore, and the majority of the images were taken within the last 5 years. 
Some examples of unsuitable images are also shown in \autoref{fig:case_study:invalid_map}, such as nighttime images, images with blur, low visual complexity, and low quality, and images taken at extremely high speeds. 
We applied rather strict criteria in this case by specifying the view direction and platform; however, these criteria can be flexibly adjusted for each use case as required by users. 
This filtering approach ensures a consistent and high-quality dataset by removing potentially unsuitable images in the crowdsourced images. 

\subsection{Clustering analysis}
For the clustering analysis, we used indicators relevant to urban characteristics from Places365 classification (i.e., highway, residential neighborhood, construction site, tree farm, forest path, and forest road), PlacePulse classification (i.e., beautiful, boring, depressing, lively, safe, and wealthy) and semantic segmentation (i.e., sky, building, vegetation, road, and sidewalk). 
These indicators were aggregated by taking the mean values to Uber's H3 grid \citep{uber_h3_2018} at a resolution of nine. 
After the z-score standardization, we applied \textit{k}-means clustering with k=5 clusters. 
This number was chosen through a combination of quantitative and qualitative considerations: while the elbow method suggested a range of potentially optimal \textit{k} values, we selected five clusters as it provided a balance between statistical validity and human-interpretable differentiation of urban streetscapes.

\begin{figure}
    \centering
    \includegraphics[width=\linewidth]{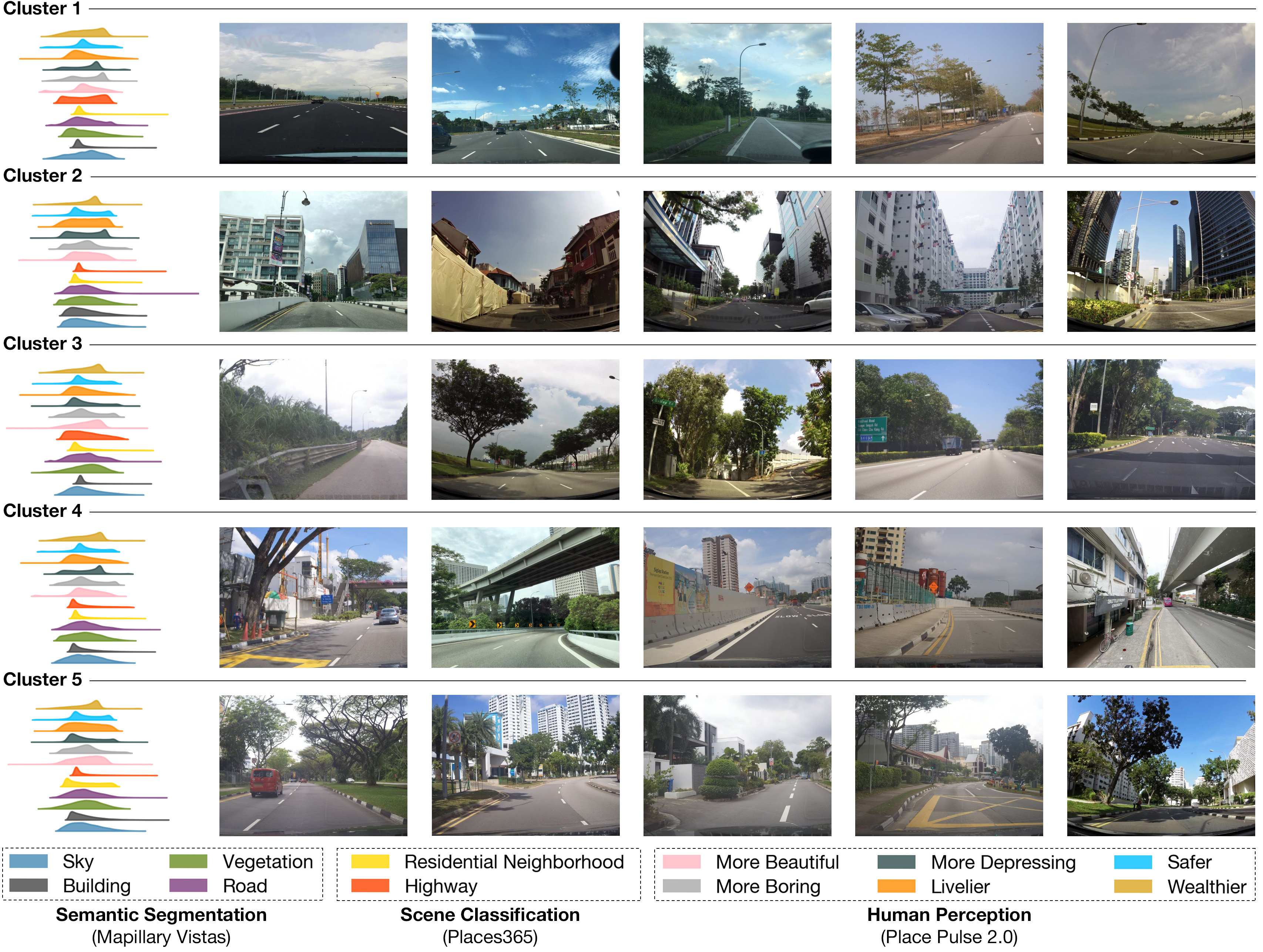}
    \caption{Representative street-level images and their corresponding feature profiles for the five visual clusters identified in Singapore. The stacked plots on the left show the relative intensity of each feature (semantic segmentation based on Mapillary Vistas, scene classification based on Places365, and human perception based on PlacePulse) for each cluster. Cluster 1 represents open highways, Cluster 2 shows dense urban areas, Cluster 3 captures green highways, Cluster 4 represents mixed urban infrastructure, and Cluster 5 shows residential areas with mature vegetation. Source of imagery: Mapillary.}
    \label{fig:case_study:visual_clusters_plot}
\end{figure}

\begin{figure}
    \centering
    \includegraphics[width=\linewidth]{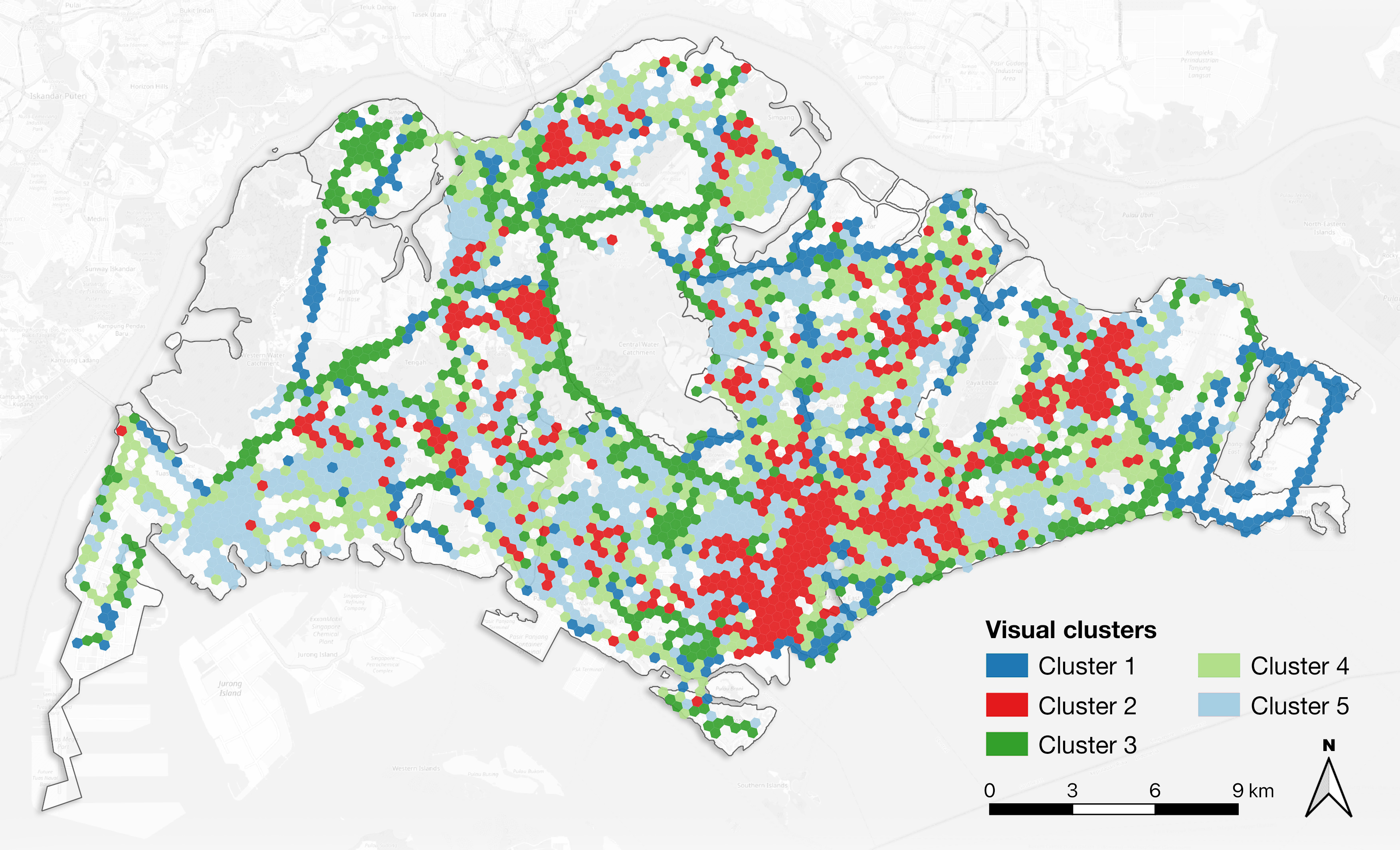}
    \caption{Spatial distribution of the five visual clusters across Singapore. The clusters show distinct spatial patterns: Cluster 1 (blue) primarily follows major highways, Cluster 2 (red) concentrates in dense urban areas and town centers, Cluster 3 (green) follows green corridors, while Clusters 4 (light green) and 5 (light blue) represent different types of residential areas distributed throughout the city.}
    \label{fig:case_study:visual_clusters}
\end{figure}

\autoref{fig:case_study:visual_clusters_plot} illustrates the five clusters formed from \textit{k}-means clustering. 
The ridge plots on the leftmost column display distributions of some relevant indicators in the respective clusters and the images on the right side of the ridge plots are randomly selected examples of SVI for the clusters. 
As shown in the sample images, Cluster 1 represents open highways with high sky view factors, probability of being classified as highway, as well as depressingness and boringness scores, and low green view index as well as safety and liveliness scores. 
Cluster 2 includes dense urban areas, characterized by tall buildings, urban infrastructure, and commercial developments. 
Cluster 3 is distinguished by green highways and corridors with abundant roadside vegetation. 
Cluster 4 shows mixed urban infrastructure with medium-rise buildings and moderate vegetation, while Cluster 5 represents residential neighborhoods with mature tree cover and lower building density. 
The spatial distribution of these clusters across Singapore in \autoref{fig:case_study:visual_clusters} reveals clear patterns, with Cluster 1 (blue) following major highway networks, Cluster 2 (red) concentrated in urban centers, Cluster 3 (green) along green corridors, and Clusters 4 and 5 (light green and light blue respectively) distributed across residential areas throughout the city.

\section{Extensibility and export} \label{sec:extensibility_export}
\subsection{Extending ZenSVI}
Our package is designed in a way that allows future development and extensions that suit a specific context such as a data stream or application. 
For example, the download sub-package can be extended further by including platforms other than Mapillary and KartaView. 
To demonstrate this extensibility, we added another source of SVI data --- an official SVI platform supported by the City of Amsterdam\footnote{\url{https://amsterdam.github.io/projects/open-panorama/}}. 
Other commercial platforms, crowdsourced services, and city-led initiatives can be similarly incorporated into ZenSVI in the future. 

The computer vision sub-package can be extended with new models, such as zero-shot segmentation \citep{kirillov_segment_2023} and various generative models for text-to-image \citep{saharia_photorealistic_2022}, image captioning \citep{li_blip_2022}, image-to-sound \citep{zhuang_hearing_2024}, image-to-video \citep{singer_makeavideo_2022}, image-to-image \citep{ito_translating_2024}, and inpainting \citep{lugmayr_repaint_2022}. 
The visualization sub-package can also be expanded with additional plot types like alluvial plots for time-series analysis. 
While leveraging ecosystems like HuggingFace, ZenSVI provides an abstraction layer that simplifies data handling while supporting models from multiple platforms through a consistent interface.

Lastly, ZenSVI's software components are released under the MIT license, which grants users extensive freedom to use, modify, distribute, and sell the software without significant restrictions. 
This permissive license allows users to modify and incorporate the code into their projects, whether open-source or proprietary, with only minimal requirements of preserving the copyright notice and license terms. 
However, it is important to note that the utilization of ZenSVI in conjunction with Mapillary or KartaView datasets necessitates adherence to their respective Creative Commons Attribution-ShareAlike 4.0 International (CC BY-SA 4.0) licensing protocols. 
These protocols mandate appropriate attribution and stipulate that any derivative works must be distributed under identical licensing conditions.

\subsection{Import and export of the data, integration, and downstream analyses thanks to ZenSVI}
The modularized structure of ZenSVI, with its independent download and other processing sub-packages, allows users to flexibly import and export data that work with external data sources and software tools.
For example, users can apply ZenSVI's transformation and computer vision sub-packages to SVI from other sources not covered in the current version, such as Baidu Street View and Bing Maps.
Most of the functions in ZenSVI also allow users to save outputs directly to external files. 
Thus, it is easy to integrate the analysis in our package with other packages. 
For example, users can download the SVI and perform semantic segmentation to compute the green view index, which can be saved to a CSV file. 
This design enables users to perform subsequent analysis and visualization in external software. 
For example, one can conduct a statistical analysis of the index in R and visualize the results in QGIS. 
Such a bridge facilitates conducting analyses and visualization in their most familiar environment while efficiently conducting the SVI analysis in ZenSVI. 

\section{Conclusion} \label{sec:discussion_conclusion}
SVI has gained popularity in many research domains for its scalability, spatial resolution, and capability to capture street-level information with high veracity and usability. 
Despite many applications demonstrated in the literature, the lack of standardized, cohesive, and reproducible software makes it difficult for other researchers to reproduce or expand existing methodologies that rely on such data and develop new ones. 
To fill this gap, we designed and implemented ZenSVI --- an integrated Python package that incorporates a comprehensive set of tools, simplifying the process of SVI analysis and contributing to the acceleration of research relying on this emerging and powerful geospatial dataset. 

The introduction of ZenSVI significantly advances the principles of open science and enhances research reproducibility in urban analytics. 
By offering a unified open-source solution, ZenSVI reduces the barrier to entry for researchers who may lack advanced technical skills, thus democratizing access to SVI analysis. 
The extensive documentation ensures that users can effectively replicate studies and build upon existing research, while the standardized workflow facilitates comparative studies across different contexts and regions.

The integration of diverse functionalities into a one-stop Python package enables users to conduct end-to-end SVI analysis with minimal code. 
For example, users can analyze the perception of safety in areas like \texttt{Fifth Avenue, New York} by downloading SVIs, converting panoramic images to 90-degree field-of-view perspectives using \texttt{transform\_images}, calculating image greenness and openness via the \texttt{segment} method, and determining safety perception scores using \texttt{calculate\_score}. 
This streamlined process demonstrates ZenSVI's capability to facilitate detailed urban studies and support informed decision-making.

Moving forward, ZenSVI's extensible architecture allows for continuous enhancement through new features and models. 
Future developments will focus on implementing text-to-image generative models and multi-modal translation capabilities, as well as expanding support for satellite, aerial, and CCTV imagery analysis. 
We also plan to create ready-to-use open datasets of SVI-derived indicators globally, further supporting users without extensive technical experience. 
Additionally, ZenSVI will serve as a platform for researchers to publish and share novel models, fostering collaboration and innovation in urban analysis and related fields.

\section*{Acknowledgments}
This research was funded by the Singapore International Graduate Award (SINGA) scholarship provided by the Agency for Science, Technology, and Research (A*STAR), the NUS Graduate Research Scholarship, and the President’s Graduate Fellowship, all granted by the National University of Singapore (NUS).
This research has been supported by Takenaka Corporation.
This research has been supported by the Research Activities Fund of City University of Hong Kong.
This research is part of the project Large-scale 3D Geospatial Data for Urban Analytics, which is supported by the National University of Singapore under the Start Up Grant R-295-000-171-133.
This research is part of the project Multi-scale Digital Twins for the Urban Environment: From Heartbeats to Cities, which is supported by the Singapore Ministry of Education Academic Research Fund Tier 1.
The research was partially conducted at the Future Cities Lab Global at the Singapore-ETH Centre, which was established collaboratively between ETH Z\"urich and the National Research Foundation Singapore (NRF) under its Campus for Research Excellence and Technological Enterprise (CREATE) programme.
We express our gratitude to the members of the NUS Urban Analytics Lab for their valuable discussions and insights.
We would like to thank the developers of the open-source software packages that made ZenSVI possible.
We also acknowledge the contributors of OpenStreetMap, Mapillary, and KartaView, and other platforms, for providing valuable open data resources that support street-level imagery research and applications.

\section*{Declaration of generative AI and AI-assisted technologies in the writing process}
During the preparation of this work, the author(s) used ChatGPT in order to format tables and proofread. After using this tool/service, the author(s) reviewed and edited the content as needed and take(s) full responsibility for the content of the published article.

\end{document}